\documentclass[journal,comsoc]{IEEEtran}
\usepackage[T1]{fontenc}

\usepackage{algorithm}
\usepackage{algorithmic}
\usepackage[algo2e]{algorithm2e}

\usepackage{hhline}
\usepackage{fixltx2e} 
\usepackage{booktabs}
\usepackage{xcolor}
\usepackage{hyperref}
\usepackage{setspace}

\usepackage[short]{optidef}
\usepackage{amsfonts}
\usepackage{amsthm}
\usepackage[utf8]{inputenc}
\usepackage{subcaption}
\usepackage{lipsum}
\usepackage{multirow}
\usepackage{arydshln}
\usepackage{filecontents,catchfile}
\usepackage{amsmath}
\usepackage{dblfloatfix}
\usepackage{float}

\newcommand{\B}{\boldsymbol}
\newcommand{\BM}{\mathbf}
\newcommand{\M}{\mathrm}

\newcommand{\Cc}{C_\textsf{c}}
\newcommand{\Cs}{C_\textsf{s}}
\newcommand{\Loss}{\mathcal{L}}%{\textsf{Loss}}
\newcommand{\Norma}{\textsf{Normalized}}
\newcommand{\tr}{\textsf{tr}}

\newcommand{\net}{\text{N}}
\newcommand{\Pre}{_\textsf{Pre}}
\newcommand{\Post}{_\textsf{Post}}
\newcommand{\Comp}{\textsf{Compression}}

\usepackage[cmintegrals]{newtxmath}

\begin{document}

\title{Corella: A Private Multi Server Learning Approach\\ based on Correlated Queries}

\author{Hamidreza~Ehteram,
	Mohammad~Ali~Maddah-Ali,
	and~Mahtab~Mirmohseni% <-this % stops a space
	\thanks{Hamidreza~Ehteram and Mahtab~Mirmohseni are with the Department of Electrical Engineering, Sharif University of Technology, Tehran, Iran (emails: ehteram.hamidreza@ee.sharif.edu, mirmohseni@sharif.edu).}% <-this % stops a space
	\thanks{Mohammad~Ali~Maddah-Ali is with Nokia Bell Labs, Holmdel, NJ, USA (email: mohammad.maddahali@nokia-bell-labs.com).}% <-this % stops a space
}

\maketitle

\begin{abstract}
The emerging applications of machine learning algorithms on mobile devices motivate us to offload the computation tasks of training a model or deploying a trained one to the cloud or at the edge of the network. One of the major challenges in this setup is to guarantee the privacy of the client data. Various methods have been proposed to protect privacy in the literature. Those include (i) adding noise to the client data, which reduces the accuracy of the result, (ii) using secure multiparty computation (MPC), which requires significant communication among the computing nodes or with the client, (iii) relying on homomorphic encryption (HE) methods, which significantly increases computation load at the servers. In this paper, we propose \emph{Corella} as an alternative approach to protect the privacy of data. The proposed scheme relies on a cluster of servers, where at most $T \in \mathbb{N}$ of them  may collude, each running a learning model (e.g., a deep neural network). Each server is fed with the client data, added with \emph{strong} noise, independent from user data. The variance of the noise is set to be large enough to make the information leakage to any subset of up to $T$ servers information-theoretically negligible. On the other hand, the added noises for different servers are \emph{correlated}. This correlation among the queries allows the parameters of the models running on different servers to be \emph{trained} such that the client can mitigate the contribution of the noises by combining the outputs of the servers, and recover the final result with high accuracy and  with a minor computational effort. In other words, instead of running a learning algorithm on top of a privacy-preserving platform, enabled by MPC or HE, in the proposed approach, the model is trained in presence of the strong noises, correlated across the servers, to achieve high accuracy, without communication and computation overheads of MPC or HE. We evaluate Corella framework for two learning algorithms, the classification, using deep neural networks, and the autoencoder, as supervised and unsupervised learning tasks, respectively. Simulation results for various datasets demonstrate the accuracy of the proposed approach for both problems.
\end{abstract}

\begin{IEEEkeywords}
Privacy Preserving Machine Learning, Private Edge Computing, Secure Machine Learning.
\end{IEEEkeywords}

\IEEEpeerreviewmaketitle

\section{Introduction} \label{Introduction}

With the expansion of machine learning (ML) applications, dealing with high dimensional datasets and models, particularly for the low resource devices (e.g., mobile units), it is inevitable to offload heavy computation and storage tasks to the cloud servers. This raises a list of challenges such as communication overhead, delay, operation cost, etc. One of the major concerns, which is becoming increasingly important, is maintaining the privacy of the used datasets, either \emph{the training dataset} or \emph{the user dataset}, such that the level of information leakage to the cloud servers is under control.

The information leakage of the training dataset may occur either during \textit{the training phase} or from \textit{the trained model}. In \emph{the training phase privacy}, a data owner that offloads the task of training to some untrusted servers is concerned about the privacy of his sampled data. In \emph{the trained model privacy}, the concern is to prevent the trained model to expose information about the training dataset. On the other hand, in \emph{the user data privacy}, a client wishes to employ some server(s) to run an already trained model on his dataset while preserving the privacy of his dataset against curiosity of servers. 

There are various techniques to provide privacy in machine learning, with the following three major categories: 

\begin{enumerate}
	\item {\textbf{Randomization, Perturbation, and Adding Noise:}}
	Applying these techniques to the client data or the ML model can be used to confuse the servers and to reduce the level of information leakage, at the cost of sacrificing the accuracy of the results. 
	
	In \cite{ModelInversion,MembershipInference}, it is shown that parameters of a trained model can leak sensitive information about the training dataset. Various approaches based on the concept of differential privacy~\cite{DifferentialPrivacyF1,DifferentialPrivacyF2,DifferentialPrivacyF3} have been proposed to provide privacy to reduce this leakage. A randomized algorithm is differentially private if its output distributions for any two input adjacent datasets, i.e., two datasets that differ only in one element,  are close enough. 
	This technique has been applied to principal component analysis (PCA) \cite{dppca1,pcaDP}, support vector machines (SVM) \cite{dpsvm}, linear and logistic regression  \cite{dplr1,dplr2}, deep neural networks (DNNs) \cite{deep1DF,deep2DF}, and distributed and federated learning \cite{disSGD,MPCDP}. 
	The caveat is that the privacy-accuracy tradeoff in these approaches \cite{DPandInf} bounds the scale of randomization and thus limits its ability to preserve privacy.

	\textit{K-anonymity} \cite{KAnonymity1,KAnonymity2,KAnonymity3} is another privacy preserving framework, in which the data items, related to one individual cannot be distinguished from the data items of at least $K-1$ other individuals in the released data. It is known that $K$-anonymity framework would not guarantee a reasonable privacy, particularly for high-dimensional data \cite{kanonnew1,kanonnew2}.

	\item {\textbf{Secure Multiparty Computation:}}
	This approach exploits the existence of a cluster of servers, some of them non-colluding, to guarantee information-theoretic privacy in some classes of computation tasks like the polynomial functions~\cite{MPCs1,MPCBGW,SecShamir}. This approach can be applied to protect the privacy of the data, while executing an ML algorithm on that cluster \cite{MPCs4,MPCs5,MPCs6}. In other words, MPC algorithms have been mathematically designed such that, the computation is done accurately, while the privacy is guaranteed perfectly. On top of this private and accurate layer of computation, any algorithm  can be executed, including ML algorithms, which can be represented (or approximated) with some polynomial computation in each round.  The shortcoming of this solution is that it costs the network a huge communication overhead, that scales with the degree of the polynomials. To reduce this communication overhead, one solution is to judiciously approximate the functions with some lower-degree polynomials such that the number of interaction among the servers is reduced to some extent~\cite{mpc1}. Another solution is to rely on Lagrange coding to develop an MPC with no communication among servers \cite{Codedprivateml}. That approach can also exploit the gain of parallel computation. The major disadvantage of \cite{Codedprivateml} is that this time the number of required servers scales with the degree of the polynomials.

	\item {\textbf{Homomorphic Encryption:}} 
	Homomorphic encryption~\cite{HE3} is another cryptography tool that can be applied for ML applications \cite{HE4,HE5,HE6,HE7}. It creates a cryptographically secure framework between the client and the servers, that allows the untrusted servers to process the encrypted data directly. However computation overhead of HE schemes is very high. This disadvantage is directly reflected in the time needed to train a model or use a trained one as reported in \cite{HE1,HE2}. 
	On the other hand, this framework is based on computational hardness assumption, and does not guarantee information-theoretic privacy.
\end{enumerate}

\subsection{Our contributions}

In this paper, we propose Corella as an alternative approach to preserve privacy in offloading ML algorithms, based on sending  \emph{correlated queries} to the  servers.  These correlated queries are generated by adding strong noises to the user data, where the added noises are independent of the data itself, but correlated across the servers. The variance of the noises is large, which makes the information leakage to the servers negligible. On the other hand, the correlation among the noises allows us to train the parameters of the model such that the user can mitigate the contribution of the noises and recover the result by combining the answers received from the servers.  Note that in this method each server runs a regular machine learning model (say a deep neural network) with no computation overhead. In addition, there is absolutely no communication among the servers. Indeed, servers may not even be aware of the existence of each other.  Thus the proposed scheme provides information-theoretic privacy,  while maintaining the communication and computation costs affordable, and still achieves high accuracy. 

In particular, we assume a client with limited computation and storage resources who wishes to run an ML task on his data. Thus for processing, it relies on a cluster of $N \in \mathbb{N}$ servers. We consider a semi-honest setup, meaning that all servers are honest in compliance with the protocol,  but an arbitrary subset of up to $T \in \mathbb{N}$ of them may collude to learn about data. Corella framework is designed for this setup by feeding correlated noisy data to each server. The added noises are designed such that the information leakage to any $T$ colluding servers remains negligible, while the system can be trained to achieve high accuracy. We apply Corella for both supervised and unsupervised learning. 

In summary, Corella offers the following desirable features: 
\begin{enumerate}
	\item Thanks to the designed strong noise added to the user data, the information leakage to any subset of up to $T$ servers is negligible.
	\item The correlation among the noises enables the system to be trained such that the user can mitigate the contribution of the noises by combining the correlated answers and recover the final result with high accuracy. 
	\item  There is no communication among the servers. Indeed, the servers may not be aware of the existence of the others.  Moreover, the computation load per server is the same as running a classical learning algorithm, e.g., a deep neural network.
\end{enumerate}

In other words, the proposed scheme resolves the tension between privacy and accuracy, and still avoids the huge computation and communication load (or employing a large number of servers), needed in HE and MPC schemes. This is particularly important, when communication resources are scarce, e.g., wireless networks.

%\pagebreak 

{\bf Remark~1:} Note that in the schemes based on randomization, perturbation, and adding noise, there is no opportunity to effectively eliminate the contribution of those privacy-protective actions (i.e., the added confusion). Thus, increasing the strength of those actions, e.g., noise variance, to improve the privacy will deteriorate the level of accuracy. This challenge has been resolved in Corella, by adding correlated noises for different servers. This provides the opportunity for the system to mitigate the contribution of the noises by combining the output of the servers. Indeed, the system gradually learns to exploit this opportunity during the training phase. However, in conventional schemes  based on randomization, perturbation, and adding noise, this opportunity does not basically exist to be exploited.

On the other hand, in MPC based schemes,  the noises added for different servers are correlated. This is similar to what we do in Corella. 
However, in MPC, the algorithm is developed to completely eliminate the added noises. In other words, unlike Corella, the system does not learn to mitigate the effect of the noises, but has been mathematically designed to absolutely remove them. This is done at the cost of prohibitive communication load among the servers, or using extra servers,   proportional to the degree of the function for  polynomial models in both cases.  However, in Corella, the model, including  the one deployed in the servers and pre- and post-processing ones in the client, is gradually trained to mitigate the contribution of the noises. This means that in the initial iterations of the learning phase, the effect of the noises is not taken care of, however, thorough learning,  the effect of the noises will be mitigated. This is done without any communication among the servers and with very few servers. 

The rest of the paper is organized as follows.  Section~\ref{ProblemSettingSection} formally presents Corella framework. Section~\ref{MethodSection} is indicated to explaining how to use Corella framework for two popular learning tasks. In Section~\ref{ExperimentsSection}, the experimental results are discussed. Section~\ref{ConclusionSection} is dedicated to concluding remarks.

\subsection {Notations} \label{Notationssubsec}
Capital italic bold letter $\B{X}$ denotes a random vector. Capital non-italic bold letter $\BM{X}$ denotes a random matrix. Capital non-italic non-bold letter $\M{X}$ denotes a deterministic matrix. $I(\B{X};\B{Y})$ indicates the mutual information between the two random vectors $\B{X}$ and $\B{Y}$. 
For a function $g$, the computation cost (e.g., the number of multiplications) and storage cost (e.g., the number of parameters) are denoted as $\Cc (g)$ and $\Cs (g)$, respectively. $\Norma(X)$, for $X=[x_1,\ldots,x_n]^\top \in \mathbb{R}^n$, is defined as $\frac{X-\mu}{\sigma}$, where $\mu = \frac{1}{n} \sum_{i=1}^n x_i$ and $\sigma^2 = \frac{1}{n} \sum_{i=1}^n (x_i-\mu)^2$. $\log x$ is calculated in base $2$ (i.e., $\log_2 x$). For $T \in \mathbb{N}$, $[T]=\{1,\ldots,T\}$. $\mathcal{N}(\mu,\sigma^2)$ denotes Gaussian distribution with mean $\mu$ and variance $\sigma^2$. $\M{W}[:,\{t_1,\ldots,t_T\}]$ denotes a submatrix of $\M{W}$, consisting of the columns $t_1,\ldots,t_T$ of matrix $\M{W}$.

\section{General Corella Framework} \label{ProblemSettingSection}

We consider a system including a client, with limited computation and storage resources, and $N$ servers. The client has an individual data and wishes to run an ML algorithm (e.g., deep neural networks, autoencoder) on it with the aid of the servers, while he wishes to keep his data private from the servers. 
All servers are honest, except up to $T<N$ of them, which are semi-honest. It means that all servers follow the protocol, but an arbitrary subset of up to $T$ servers may collude to gain information about the client data. 
The client sends queries to the servers, and then by combining the received answers, he will derive the result. In other words, we are concerned about privacy of the client data.

The system is operated in two phases, the training phase, and then the inference phase. 

In the training phase, the dataset $\mathcal{S}_m=\{(\B{X}^{(1)},\B{Y}^{(1)}),\ldots,(\B{X}^{(m)},\B{Y}^{(m)})\}$ consisting of $m \in \mathbb{N}$ samples are used by the client to train the model, where $(\B{X}^{(i)},\B{Y}^{(i)})$ denotes the data sample and its target (e.g., label in the supervised learning), for $i=1,\ldots,m$.
In addition, the client generates $m$ independent and identically distributed (i.i.d.) noise samples $\mathcal{Z}_m=\{(\B{Z}_1^{(1)}, \ldots, \B{Z}_N^{(1)}),\ldots, (\B{Z}_1^{(m)}, \ldots, \B{Z}_N^{(m)})\}$, where each noise sample $(\B{Z}_1^{(i)},\ldots, \B{Z}_N^{(i)})$, with $N$ \emph{correlated} components, is sampled from a joint distribution $\mathbb{P}_{\BM{Z}} = \mathbb{P}_{\B{Z}_1,\ldots, \B{Z}_N}$. 
The noise components are independent of dataset $\mathcal{S}_m$.

\begin{figure}[!t]%icv:[btp] %[ht]
	%\vskip -0.2in
	\begin{center}
		\centerline{\includegraphics[width=\columnwidth,height=8cm,keepaspectratio]{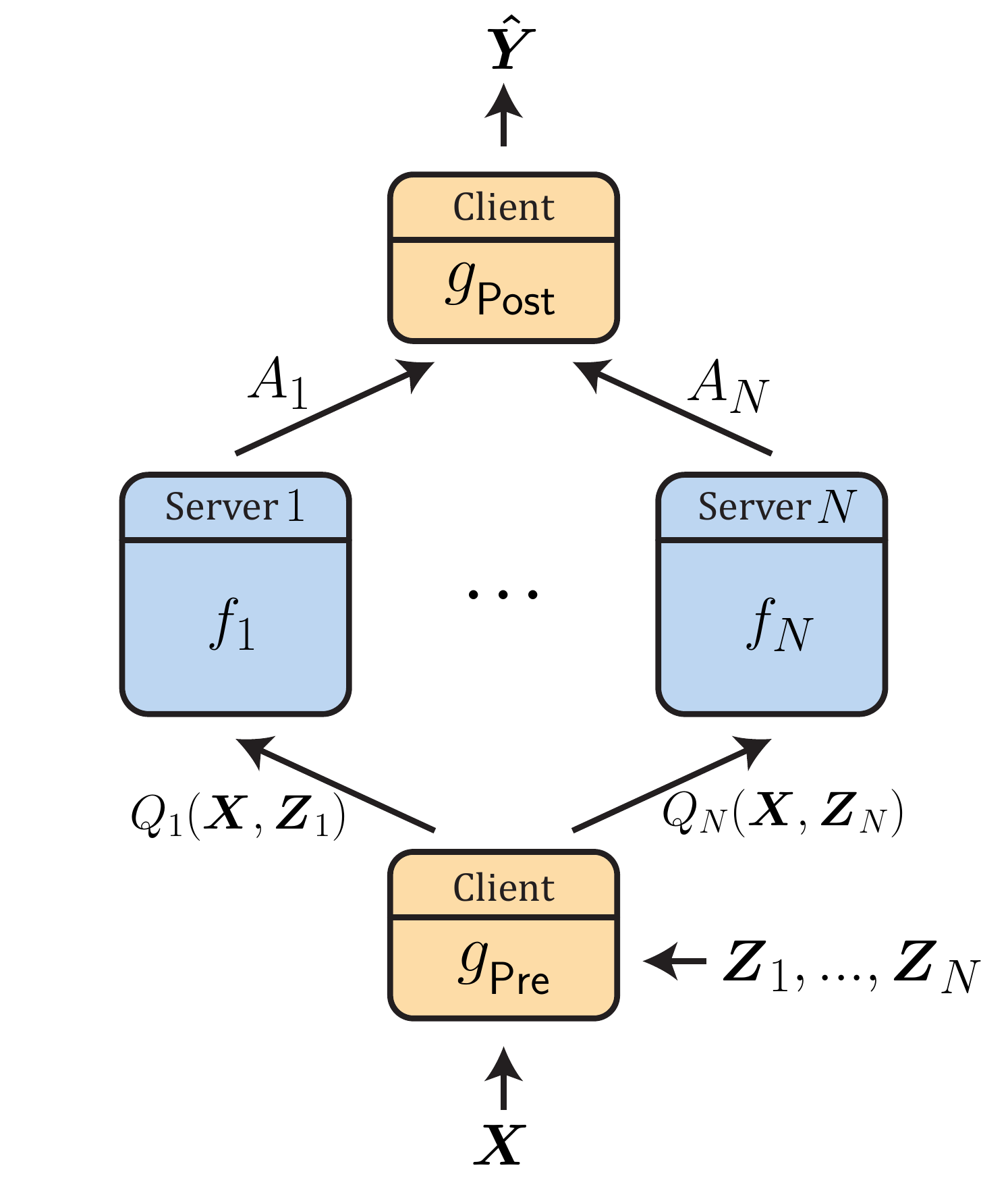}}
		\caption{The general Corella framework for privacy preserving ML}
		\label{fig:Probset}
	\end{center}
	%\vskip -0.4in
\end{figure}

The data flow diagram is shown in Fig.~\ref{fig:Probset}, where for simplicity, the sample index $(i)$ is omitted from the variables. For $i=1\ldots m$, the client, having access to dataset $\mathcal{S}_m$ and noise component set $\mathcal{Z}_m$, uses a preprocessing function $g_{\Pre}$ to generate $N$ queries 
\begin{align}
\big(Q_1(\B{X}^{(i)},\B{Z}^{(i)}_1),\ldots, Q_N(\B{X}^{(i)},\B{Z}^{(i)}_N)\big)= g_{\Pre}(\B{X}^{(i)},\B{Z}^{(i)}),
\end{align}
and sends $Q_j(\B{X}^{(i)},\B{Z}^{(i)}_j)$ to the $j$-th server, for $j=1,\ldots,N$. 
In response, the $j$-th server applies a function (a model) $f_j$, which will be trained, and generates the answer $A^{(i)}_j$ as,
\begin{align}
A^{(i)}_j = f_j(Q_j(\B{X}^{(i)},\B{Z}^{(i)}_j)).
\end{align}
By combining all answers from $N$ servers using a post-processing function $g_{\Post}$, the client estimates the target,
\begin{align}
\B{\hat{Y}}^{(i)}= g_{\Post}(A^{(i)}_1,\ldots,A^{(i)}_N),
\end{align}
while the information leakage from the set of queries to any arbitrary $T$ servers must be negligible.

In the training phase, the goal is to design or train the set of functions $\mathcal{F}=\{g_{\Pre},g_{\Post},f_1,\ldots,f_N\}$ and $\mathbb{P}_{\BM{Z}}$ according to the following optimization problem

\begin{mini}
	{\mathcal{F},\mathbb{P}_{\BM{Z}}}{\frac{1}{m} \sum\limits_{i=1}^{m} \Loss\{\B{\hat{Y}}^{(i)},\B{Y}^{(i)}\}}
	{\label{eq:probset1}}{}
	\addConstraint{I\Big(\B{X}^{(i)};\big\{Q_j(\B{X}^{(i)},\B{Z}^{(i)}_j), j \in \mathcal{T} \big\}\Big) \leq \varepsilon}
	\addConstraint{\forall i \in [m]}
	\addConstraint{\forall \mathcal{T} \subset [N], |\mathcal{T}| \le T}
\end{mini}
where $\Loss\{\B{\hat{Y}}^{(i)},\B{Y}^{(i)}\}$ shows the loss function between $\B{\hat{Y}}^{(i)}$ and $\B{Y}^{(i)}$, for some loss function $\Loss$ and the constraint guarantees to preserve $\varepsilon$-privacy in terms of information leakage through any set of $T$ queries, for some privacy parameter $\varepsilon \in \mathbb{R}_{\geq 0}$. We desire that the computation and storage costs of $g_{\Pre}$ and $g_{\Post}$ be low.

In the inference phase, to deploy this model to estimate the target of a new input $\B{X}$, the client chooses $(\B{Z}_1,\ldots,\B{Z}_N)$, sampled from designed distribution $\mathbb{P}_{\BM{Z}}$, independent of all other variables in the network, and follows the same protocol and uses the designed or trained functions set  $\mathcal{F}$.

This framework can apply to both supervised and unsupervised ML algorithms. In the next section, we show how to use Corella framework for the classification and the autoencoder tasks, as two popular examples of supervised and unsupervised learning tasks.

\textbf{Remark~2:} One of the interesting aspects of Corella framework is that the servers don't communicate with each other at all. Indeed, they may not be even aware of the existence of the other servers.

\section{Corella in Action} \label{MethodSection}

The objective in this section is to use Corella framework to develop two learning algorithms, (1) privacy-preserving classification, based on deep neural networks, and (2) privacy-preserving autoencoder, as classical examples for supervised and unsupervised learning, respectively.

\subsection{Classification Using Deep Neural Netwoks}\label{SupTasksC}
\begin{figure}[!t]%icv:[btp]%[ht]
	%\vskip -0.2in
	\begin{center}
		\centerline{\includegraphics[width=\columnwidth,height=11cm,keepaspectratio]{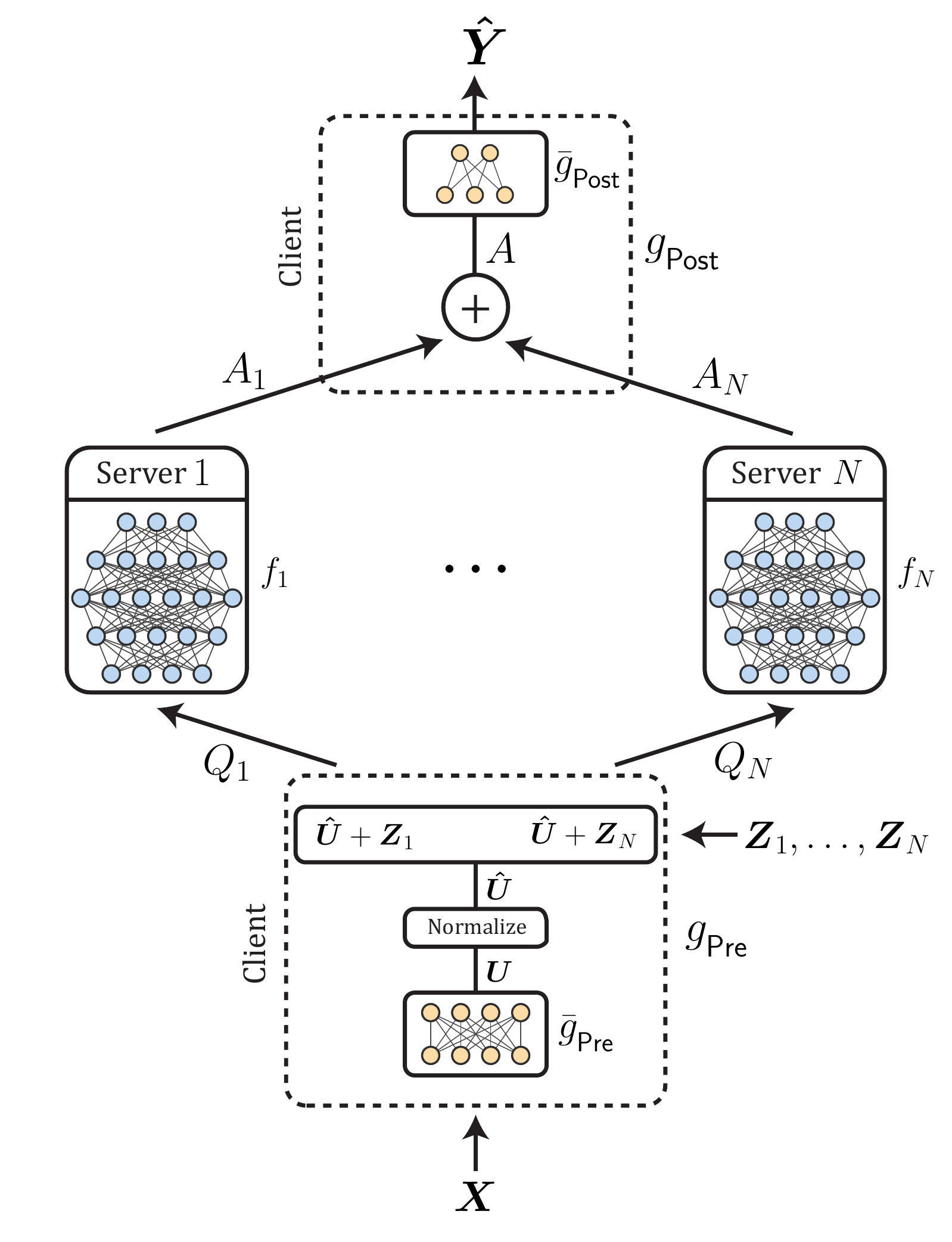}}
		\caption{Corella for classification}
		\label{fig:method}
	\end{center}
	%\vskip -0.2in
	%\vskip -0.4in
\end{figure}

In this task, the client wishes to label his data, using $N$ servers, where up to $T$ of them may collude. In this problem, $\B{Y}$ is the label of $\B{X}$. The structure of the algorithm has been shown in Fig.~\ref{fig:method}. Here, we explain the different components of the proposed algorithm.

\subsubsection{\textbf{Correlated joint distribution $\mathbb{P}_{\BM{Z}}$}} \label{partgennoise}
The following steps describe how we generate samples of $\BM{Z}$. First a matrix $\M{W} \in \mathbb{R}^{T \times N}$ is formed, such that any submatrix of size $T \times T$ of $\M{W}$ is \emph{full rank}.
Then a random matrix $\BM{\bar{Z}} = [\B{\bar{Z}}_1,\ldots,\B{\bar{Z}}_T] \in \mathbb{R}^{s \times T}$, independent of $\B{X}$, is formed where each entry is chosen independently and identically from $\mathcal{N}(0,\sigma^2_{\bar{Z}})$, where $s$ is the size of each query and $\sigma^2_{\bar{Z}}$ is a positive real number, denoting the variance of each entry. Then, let $\BM{Z}$ be 
\begin{align}
\label{def:Z}
\BM{Z} \stackrel{\footnotesize{\triangle}}{=} [\B{Z}_1, \ldots, \B{Z}_N] = \BM{\bar{Z}}  \M{W}.
\end{align}

\subsubsection{\textbf{Preprocessing function $g_{\Pre}$}}
This function, which generates the queries, consists of three blocks: (i) the sample vector $\B{X}$ is passed through a neural network with learnable parameters, denoted by $\bar{g}_{\Pre}$; (ii) the output of first block, $\B{U}=\bar{g}_{\Pre}(\B{X})$, is $\Norma$ to $\B{\hat{U}}$ (as defined in the notation \ref{Notationssubsec}); (iii) the query of the $j$-th server, $Q_j$, is generated by adding the noise component $\B{Z}_j$ to $\B{\hat{U}}$. Therefore, 
\begin{align}
\label{def:G}
Q_j = G(\B{X}) + \B{Z}_j=\Norma(\bar{g}_{\Pre}(\B{X}))+ \B{Z}_j.
\end{align}

Since the client has limited computing resources, $\bar{g}_{\Pre}$ can be a neural network with one layer or even an identity function. As there is no a priori bound on the range of the output of a neural network (e.g., $\bar{g}_{\Pre}$), the advantage of $\Norma(\cdot)$ function is that provides such a bound. 
As we will see in Subsection~\ref{subsubprpr}, a large enough noise variance $\sigma^2_{\bar{Z}}$ is sufficient to make the constraint of optimization~\eqref{eq:probset1} be satisfied, independent of the choice of $\bar{g}_{\Pre}$ function.

\subsubsection{\textbf{Post-processing function $g_{\Post}$}} 
We form $g_{\Post}$ by running a neural network with learnable parameters,  denoted by $\bar{g}_{\Post}$, over the sum of the received answers from the servers. Therefore,
\begin{align}
\label{def:A}
\B{\hat{Y}} = \bar{g}_{\Post}({A}) = \bar{g}_{\Post}\Big({\sum_{j=1}^N A_j}\Big).
\end{align} 

To limit the burden of the computation and storage on the client, we desire to keep these costs at the client side low. Hence in the implementation, we allocate at most a single-layer neural network to $\bar{g}_{\Pre}$ and $\bar{g}_{\Post}$.

\subsubsection{\textbf{Functions $f_1$ to $f_N$}} These functions are chosen as some multi-layer neural networks with learnable parameters. The parameters of $f_1$ to $f_N$ will be different.

\subsubsection{\textbf{Training}}\label{subsubtraining} To train the learnable parameters of $\mathcal{F}$, we use some a particular form of gradient descent optimization algorithms to minimize the loss of optimization~\eqref{eq:probset1}. In other words, we train a model, consisting of $N$ separate neural networks $f_1$ to $f_N$ and two networks $\bar{g}_{\Pre}$ and $\bar{g}_{\Post}$. The details of this method are presented in Algorithm \ref{alg:TestPrivacyAlg}. In this algorithm, the parameters of the model are denoted by $\theta$ and the training batch size is indicated by $b$. 

\begin{algorithm}[H]%[btp]%[tb]%[H]
	\SetAlgoLined
	\SetKwInput{Input}{input}
	\SetKwInput{Output}{output}
	
	\Input{
		$\mathcal{S}_m$,
		$\sigma_{\bar{Z}}$,
		$\M{W}$,
		$\bar{g}_{\Pre}(\cdot;\theta_{\bar{g}_{\Pre}})$,
		$\bar{g}_{\Post}(\cdot;\theta_{\bar{g}_{\Post}})$,
		$f_j(\cdot;\theta_{f_j})$,
		$b$}
	
	\Output{
		$\mathbb{P}_{\BM{Z}}$ and updated
		$\bar{g}_{\Pre}(\cdot;\theta_{\bar{g}_{\Pre}})$,
		$\bar{g}_{\Post}(\cdot;\theta_{\bar{g}_{\Post}})$,
		$f_j(\cdot;\theta_{f_j})$%; i=1,...,N$
	}
	
	\BlankLine
	
	$i \gets 1,\ldots,b$
	
	$s \gets$ the output size of $\bar{g}_{\Pre}(\cdot;\theta_{\bar{g}_{\Pre}})$
	
	\SetKwFunction{newfunction}{$\mathbb{P}_{\BM{Z}}$}
	\SetKwProg{Fn}{function}{}{end}
	
	\Fn{\newfunction}{
		Draw $s \times T$ $i.i.d.$ noise samples from $\mathcal{N}(0,\sigma_{\bar{Z}}^2)$
		
		Shape the noise samples to $s \times T$ matrix $\BM{\bar{Z}}$
		
		Compute the noise components
		
		$[\B{Z}_1, \ldots,\B{Z}_N] \gets \BM{\bar{Z}}\M{W}$

		\KwRet $(\B{Z}_1,\ldots,\B{Z}_N)$
		
	}
	
	\For{\textit{the number of training iterations}}{
		{\bfseries Forward path:}
		
		Draw $b$ minibatch samples from $\mathcal{S}_m$
		
		$\{(\B{X}^{(1)},\B{Y}^{(1)}),\ldots,(\B{X}^{(b)},\B{Y}^{(b)})\}$
		
		Draw $b$ $i.i.d.$ noise samples from $\mathbb{P}_{\BM{Z}}$
		
		$\{(\B{Z}_1^{(1)},\ldots, \B{Z}_N^{(1)}),\ldots, (\B{Z}_1^{(b)}, \ldots, \B{Z}_N^{(b)})\}$

		Compute the client features
		
		$\B{U}^{(i)} \gets \bar{g}_{\Pre}(\B{X}^{(i)};\theta_{\bar{g}_{\Pre}})$
		
		Normalize the client features
		
		$\B{\hat{U}}^{(i)} \gets \Norma(\B{U}^{(i)})$
		
		\For{$j=1,\ldots,N$}{
			Compute the query of the $j$-th server
			
			${Q}^{(i)}_j \gets \B{\hat{U}}^{(i)} + \B{Z}^{(i)}_j$%; j=1,...,b; i=1,...,N$
			
			Compute the answer of the $j$-th server
			
			${A}^{(i)}_j \gets f_j({Q}^{(i)}_j;\theta_{f_j})$
		}
		
		Compute the sum of the answers
		
		${A}^{(i)} \gets \sum_{j=1}^N {A}^{(i)}_j$
		
		Compute the client predicted labels
		
		$\B{\hat{Y}}^{(i)} \gets \bar{g}_{\Post}({A}^{(i)};\theta_{\bar{g}_{\Post}})$

		Compute the loss
		
		$L(\theta_{\bar{g}_{\Pre}},\theta_{\bar{g}_{\Post}},\theta_{f_1},\ldots,\theta_{f_N}) \gets \frac{1}{b} \sum\limits_{i} \Loss\{\hat{\B{Y}}^{(i)},\B{Y}^{(i)}\}$
		
		{\bfseries Backward path:}
		
		The backpropagation (BP) algorithm:
		
		Update $\theta_{\bar{g}_{\Post}}$ %by descending its stochastic gradient
		
		\For{$j=1,\ldots,N$}{
			
			Compute the BP of the client to the $j$-th server %$\nabla_{{A}^{(i)}_j}L$

			Update $\theta_{f_j}$ %by descending its stochastic gradient

			Compute the BP of the $j$-th server to the client
			
		}
		
		Update $\theta_{\bar{g}_{\Pre}}$ %by descending its stochastic gradient
		
	}
	
	\caption{Corella: $\varepsilon$-privacy with $N$ servers, up to $T$ colluding}
	\label{alg:TestPrivacyAlg}

\end{algorithm}

\subsubsection{\textbf{Proof of privacy preserving}}\label{subsubprpr} In Theorem~\ref{Theorem1}, we show that the proposed method satisfies $\varepsilon$-privacy, if
\begin{align}
\label{ineq:sigma}
\sigma^2_{\bar{Z}} \geq \frac{1}{2\ln 2} ps/\varepsilon,
\end{align}
where $s$ is the size of each query and
\begin{align}
\label{eq:p}
p = \max\limits_{\M{\Omega} \in \mathcal{W}} \{ \vec{1}^{\,\top}(\M{\Omega} \M{\Omega}^\top)^{-1} \vec{1} \} \in \mathbb{R}_+.
\end{align}
$\mathcal{W}$ denotes the set of all $T \times T$ submatrices of $\M{W}$ and $\vec{1}$ denotes the all-ones vector with length $T$.

\newtheorem{myTheorem}{Theorem}
\begin{myTheorem}[$\varepsilon$-privacy]\label{Theorem1}
	Let  $\B{X}$ be a random vector, sampled from some distribution $\mathbb{P}_{\B{X}}$, $\BM{Z} = [\B{Z}_1, \ldots, \B{Z}_N]$ and $\BM{Q} = [Q_1, \ldots, Q_N]$ as defined in~\eqref{def:Z} and \eqref{def:G}. If conditions~\eqref{ineq:sigma} and \eqref{eq:p} are satisfied, then for all $\mathcal{T}\subset \{1,\ldots, N\}$ of size $T$, we have 
	\begin{align}
	I (\B{X};  \{Q_j,  j \in \mathcal{T} \}) \leq \varepsilon. 
	\end{align}
	
	\begin{proof}
		Let $\M{K} = \mathbb{E}[ G(\B{X}) G(\B{X})^T]$ denote the covariance matrix of $G(\B{X})$. Since $G(\B{X})$ is $\Norma$, then  $\tr(\M{K}) = s$. In addition,  consider  the set  $\mathcal{T} = \{t_1, \ldots, t_T\}$, where  $\mathcal{T} \subset [N]$ and $|\mathcal{T}|=T$, also let $\M{\Omega}_{\mathcal{T}}= \M{W}(:, \mathcal{T})$, and $\BM{Q}_{\mathcal{T}}=[Q_{t_1}, \ldots, Q_{t_T}]$. 
		Then, we have 
		\begin{align}
		\label{eq:QQ}
		\BM{Q}_{\mathcal{T}} = [G(\B{X}), \BM{\bar{Z}}]  [\vec{1}, \M{\Omega}_{\mathcal{T}}^\top]^\top,
		\end{align}
		where $\BM{\bar{Z}}$ is defined in \ref{partgennoise}. In addition, we define  
		\begin{align}
		\label{omega}
		[\omega_1,\ldots,\omega_T]= \vec{1}^\top\M{\Omega}_{\mathcal{T}}^{-1}.
		\end{align}
		Thus we have,
		\begin{align*}
		&I (\B{X};  \{Q_j,  j \in \mathcal{T} \}) = I(\B{X};\BM{Q}_{\mathcal{T}}) \\
		& \overset{(\text{a})}{=} I(\B{X};\BM{Q}_{\mathcal{T}}\M{\Omega}_{\mathcal{T}}^{-1})
		= h(\BM{Q}_{\mathcal{T}}\M{\Omega}_{\mathcal{T}}^{-1}) - h(\BM{Q}_{\mathcal{T}}\M{\Omega}_{\mathcal{T}}^{-1} | \B{X}) \\
		& \overset{(\text{b})}{=} h(\omega_1G(\B{X})+\B{\bar{Z}}_1,\ldots,\omega_TG(\B{X})+\B{\bar{Z}}_T) - h(\BM{\bar{Z}}) \\
		& \overset{(\text{c})}{\leq} \sum_{t= 1}^T \Big(h(\omega_tG(\B{X})+\B{\bar{Z}}_{t})-h(\B{\bar{Z}}_{t})\Big) \\
		&\overset{(\text{d})}{\leq}\sum_{t = 1}^T \frac{1}{2} \Big( \log\big((2\pi e)^s \det (\omega^2_t\M{K}+\sigma_{\bar{Z}}^2\M{I}_s)\big) - \log(2\pi e \sigma_{\bar{Z}}^2)^s \Big)\\
		&\overset{(\text{e})}{\leq}\frac{1}{2} \sum_{t = 1}^T \Big( \log \big( (\frac{1}{\sigma_{\bar{Z}}^2})^s (\frac{\tr(\omega_t^2\M{K}+\sigma_{\bar{Z}}^2\M{I}_s)}{s})^s \big) \Big) \\
		&\overset{(\text{f})}{=} \frac{s}{2} \sum_{t = 1}^T \log(\frac{\omega_t^2+\sigma_{\bar{Z}}^2}{\sigma_{\bar{Z}}^2})\overset{(\text{g})}{\leq}\frac{s}{2\ln 2} \sum_{t= 1}^T \frac{\omega_t^2}{\sigma_{\bar{Z}}^2}  \overset{(\text{h})}{\leq} \varepsilon,
		\end{align*}

		where 
		(a) follows since $ \M{\Omega}_{\mathcal{T}}$ is a full rank matrix; (b) follows from~\eqref{eq:QQ}, \eqref{omega} and the fact that $\BM{\bar{Z}}$ is independent of $\B{X}$; (c) follows from inequality $h(\B{A},\B{B})\leq h(\B{A}) + h(\B{B})$ for any two random vectors $\B{A}$ and $\B{B}$, and the fact that the set of random vectors $\{\B{\bar{Z}}_1,\ldots,\B{\bar{Z}}_T\}$ is mutually independent; (d) follows because $\BM{\bar{Z}}$ is independent of $\B{X}$ and jointly Gaussian distribution maximizes the entropy of a random vector with a known covariance matrix \cite{Lemma1}; (e) follows from the fact that by considering a symmetric and positive semi-definite matrix $\M{H}  = \omega_t^2 \M{K} +\sigma_{\bar{Z}}^2\M{I}_s$ with eigenvalues $\lambda_k$, we have $\det(\M{H}) = \prod\limits_{k=1}^{s} \lambda_k$ and $\tr(\M{H}) = \sum\limits_{k=1}^{s} \lambda_k$, and therefore we obtain $\det(\M{H}) \leq (\frac{\tr(\M{H})}{s})^s$ using the inequality of arithmetic and geometric means; (f) follows since $\tr(\M{K}) = s$; (g) follows due to $\ln(x+1) \leq x$; and (h) follows from \eqref{ineq:sigma}, \eqref{eq:p},  \eqref{omega}, and  by substituting $\sum\limits_{t = 1}^T \omega_t^2 = \vec{1}^{\,\top}(\M{\Omega}_{\mathcal{T}} \M{\Omega}_{\mathcal{T}}^\top)^{-1} \vec{1} \leq  p$. 
		
	\end{proof}

\end{myTheorem}

In the next section, we will evaluate the performance of the proposed scheme for several datasets.

\subsection{Autoencoder}\label{unsupTasks}

\begin{figure*}[!t]%icv:[ht]%[ht]%[btp]%[ht]
	%\vskip 0.2in
	\centering
	\begin{subfigure}{0.32\textwidth}%{0.3\textwidth}
		\centering
		\includegraphics[width=\linewidth,height=10cm,keepaspectratio]{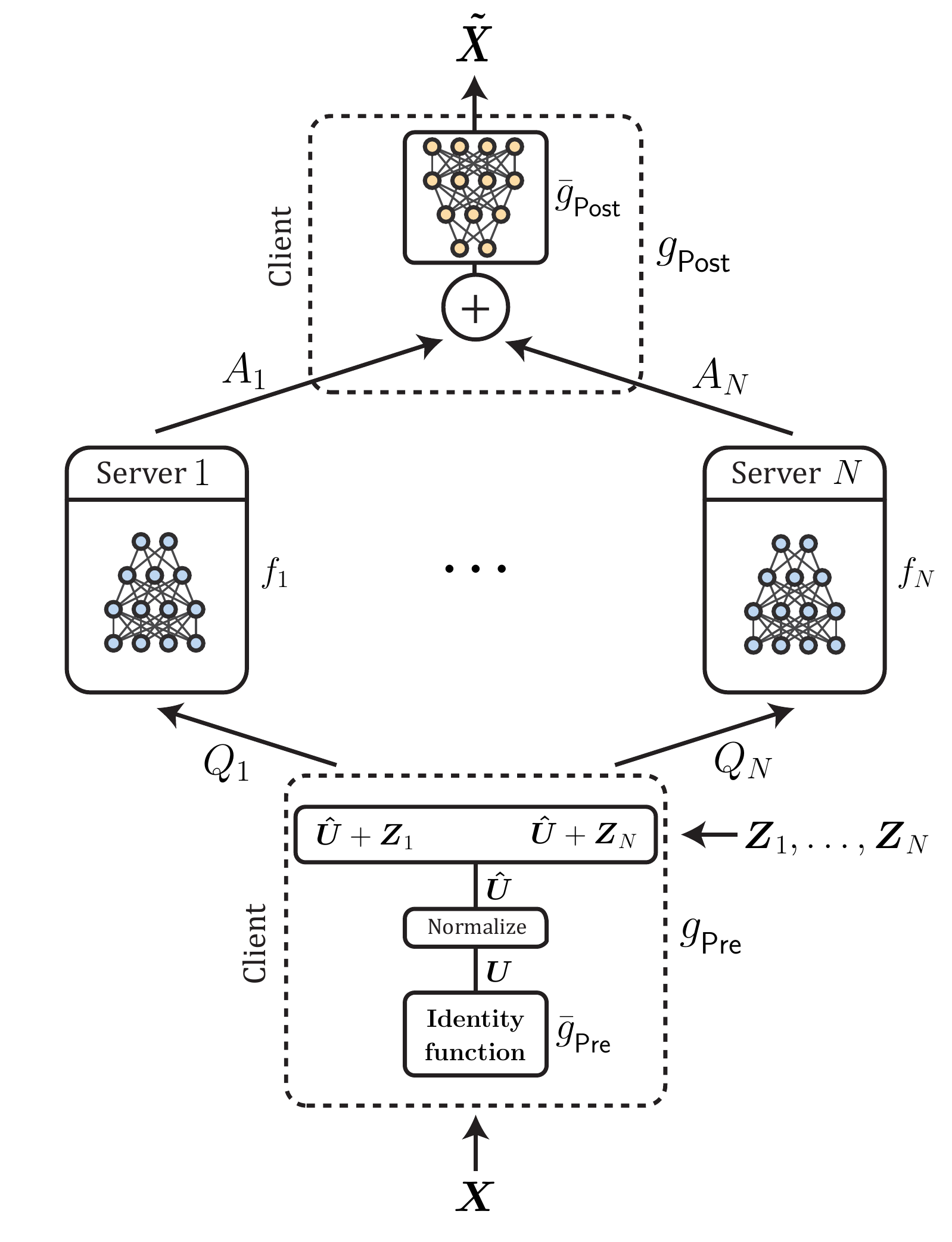}%\linewidth%\columnwidth
		\caption{Scenario 1: Offloading the encoding stage}
		\label{fig:autoen}
	\end{subfigure}%
	\hfill
	\begin{subfigure}{0.32\textwidth}
		\centering
		\includegraphics[width=\linewidth,height=10cm,keepaspectratio]{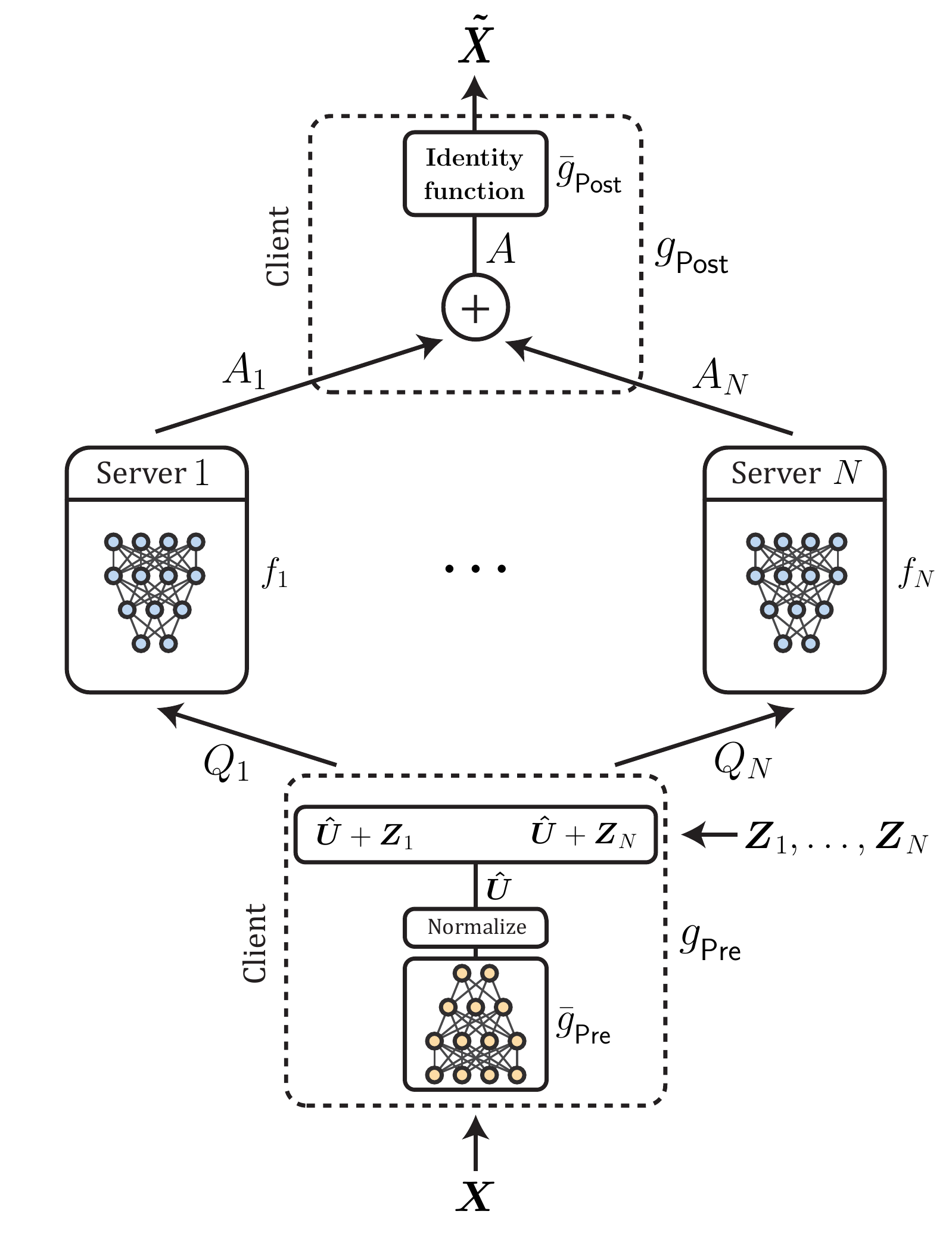}
		\caption{Scenario 2: Offloading the decoding stage}
		\label{fig:autode}
	\end{subfigure}%
	\hfill
	\begin{subfigure}{0.32\textwidth}%{0.3\textwidth}
		\centering
		%\vskip 0.3in
		\includegraphics[width=\linewidth,height=10cm,keepaspectratio]{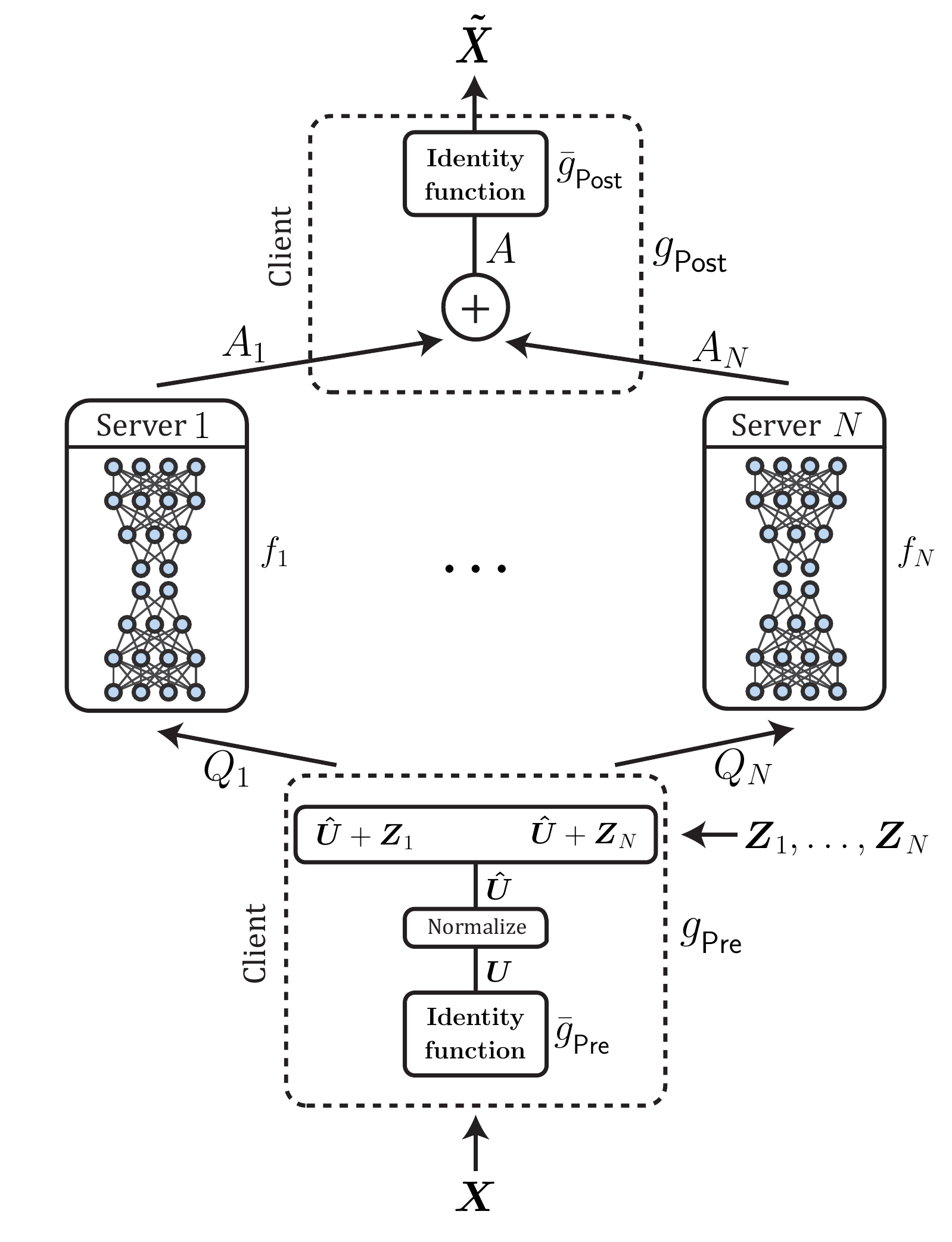}%\linewidth%\columnwidth
		%\vskip 0.1in
		\caption{Scenario 3: Offloading both encoding and decoding stages}
		\label{fig:autoed}
	\end{subfigure}%
	%\vskip -0.1in
	\caption{The method in the autoencoder task}\label{fig:automethod}
	%\vskip -0.2in
	%\vskip -0.2in
\end{figure*}

As an example of an unsupervised task, here we focus on autoencoder. An autoencoder has two stages: an encoding stage and a decoding stage. In the encoding stage, the client wishes to compress $\B{X}$ to a \textit{latent} $L_j \in \mathbb{R}^l$, and store latent $L_j$ in the $j$-th server, and at a later time, in the decoding stage, reconstruct $\B{X}$ from latent $L_j$, for $j = 1 \ldots N$. The real number $R_\Comp=\frac{x}{l}$ indicates the compression ratio, where $x$ and $l$ denote the size of $\B{X}$ and $L_j$, respectively. In this problem, $\B{Y}$ is equal to $\B{X}$. Here we explain the components of the algorithm. In terms of the process of the noise generating we follow the same steps as explained in Subsection~\ref{SupTasksC} (see \eqref{def:Z}). For other components, we have three different scenarios (shown in Fig.~\ref{fig:automethod}):

\subsubsection{\textbf{Scenario 1: Offloading The Encoding Stage} (Fig.~\ref{fig:autoen})}
In this scenario the client wishes to offload the encoding stage to the servers. To generate the queries by $g_{\Pre}$, the client passes $\B{X}$ through $\bar{g}_{\Pre}$ and then normalizes the output to $\B{\hat{U}}$. Having noise component $\B{Z}_j$, he sends query $Q_j=\B{\hat{U}}+\B{Z}_j$ to the $j$-th server. The $j$-th server applies $f_j$ to encode and compress received query $Q_j$ to latent $L_j \in \mathbb{R}^l$, and then stores $L_j$. When the client wants to reconstruct $\B{X}$, the $j$-th server sends $A_j=L_j$ to the client. The client passes the sum of received answers, i.e., $A=\sum_{j=1}^{N} A_j$, through function $\bar{g}_{\Post}$, which implements the decoding stage, to recover $\B{\tilde{X}}$ (as an estimation of $\B{X}$). In this problem, we desire that the computation and storage costs of function $\bar{g}_{\Pre}$ be low. Hence in the implementation, we may set $\bar{g}_{\Pre}$ equal to an identity function.

\subsubsection{\textbf{Scenario 2: Offloading The Decoding Stage} (Fig.~\ref{fig:autode})}
In this scenario the client wishes to offload the decoding stage to the servers. Here function $g_{\Pre}$ consists of both the encoding stage and the query generating processes, and is done at the client side. The client uses $\bar{g}_{\Pre}$ to compress $\B{X}$ to $\B{U} \in \mathbb{R}^l$, then he normalizes the output to $\B{\hat{U}}$, and sends query $Q_j=\B{\hat{U}}+\B{Z}_j$ to the $j$-th server. The $j$-th server stores latent $L_j = Q_j \in \mathbb{R}^l$. When the client wants to reconstruct $\B{X}$, the $j$-th server uses $f_j$ to decode and decompress $L_j$ to $A_j$ and send answer $A_j$ to the client. Then, the client passes $A=\sum_{j=1}^{N} A_j$ through $\bar{g}_{\Post}$ to recover $\B{\tilde{X}}$. Here, we desire that the computation and storage costs of $\bar{g}_{\Post}$ be low. Thus in the implementation, we may set $\bar{g}_{\Post}$ equal to an identity function. 

\subsubsection{\textbf{Scenario 3: Offloading Both Encoding and Decoding Stages} (Fig.~\ref{fig:autoed})}
In this scenario the client wishes to offload both encoding and decoding stages to the servers. To generate the queries by $g_{\Pre}$, the client passes $\B{X}$ through $\bar{g}_{\Pre}$ and normalizes the output to $\B{\hat{U}}$, and then sends query $Q_j=\B{\hat{U}}+\B{Z}_j$ to the $j$-th server. The $j$-th server applies $f_j^{\textsf{enc}}$ to compress the received query to latent $L_j \in \mathbb{R}^l$, and then stores $L_j$. When the client wants to reconstruct $\B{X}$, the $j$-th server applies $f_j^{\textsf{dec}}$ to decode $L_j$ to $A_j$ and send $A_j$ to the client. The client passes the sum of received answers, i.e., $A=\sum_{j=1}^{N} A_j$, through $\bar{g}_{\Post}$ to reconstruct $\B{\tilde{X}}$. In this scenario, $f_j(.)= f_j^{\textsf{dec}}(f_j^{\textsf{enc}}(.)) $ and we desire to have low computation and storage costs for $\bar{g}_{\Pre}$ and $\bar{g}_{\Post}$. Hence in the implementation, we may set both $\bar{g}_{\Pre}$ and $\bar{g}_{\Post}$ equal to an identity function.

Model training and poof of privacy-preserving in this algorithm are the same as discussed in Subsection~\ref{SupTasksC} (see \ref{subsubtraining} and \ref{subsubprpr}).

\section{Experiments}\label{ExperimentsSection}
This section is dedicated to reporting simulation results. In Subsection~\ref{subIm}, the implementation details of the proposed method is presented. In Subsections~\ref{subCla} and \ref{subAut}, the performance of the proposed method is evaluated for the classification and the autoencoder tasks, respectively.

\subsection{The Implementation Details} \label{subIm}

\begin{table*}[!t]%icv:%[t]
	\caption{Network structure. $\Norma (\cdot)$ function, normalizes its input as defined in the notation \ref{Notationssubsec}. Conv2d parameters represent the number of the input channels, the number of the output channels, the kernel size, the stride, and the padding of a 2D convolutional layer, respectively. FC parameters represent the number of the input neurons and the number of the output neurons of a fully connected layer. BatchNorm2d and BatchNorm1d parameters represent the number of the input channels and the number of the input neurons respectively for batch normalization layer.}
	\label{tab:tabStruc}
	%\vskip 0.15in
	\begin{center}
		\begin{small}
			%\begin{sc}
			\centering
			\begin{tabular}{lllll} 
				\toprule
				& \multicolumn{1}{c}{$\B{\net(\mathsf{Iden.}\rightarrow\mathsf{Iden.})}$ }                                                                                                                                                                              & \multicolumn{1}{c}{$\B{\net(\mathsf{Iden.}\rightarrow n_\text{o})}$ }                                                                                                                                                                                           & \multicolumn{1}{c}{$\B{\net(n_\text{i}\rightarrow\mathsf{Iden.})}$ }                                                                                                                                       & \multicolumn{1}{c}{$\B{\net(n_\text{i}\rightarrow n_\text{o})}$ }                                                                                                                                                     \\ 
				\toprule
				$\B{\bar{g}_{\Pre}}$   & Identity function                                                                                                                                                                                                                                     & Identity function                                                                                                                                                                                                                                               & \begin{tabular}[c]{@{}l@{}}Conv2d ($c_i$,$n_\text{i}$,(5,5),3,0)\\BatchNorm2d ($n_\text{i}$)\\ReLU \end{tabular}                                                                                           & \begin{tabular}[c]{@{}l@{}}Conv2d ($c_i$,$n_\text{i}$,(5,5),3,0)\\BatchNorm2d ($n_\text{i}$)\\ReLU \end{tabular}                                                                                                      \\ 
				\midrule
				$\B{f_j}$              & \begin{tabular}[c]{@{}l@{}}$\textsf{Normalized}$ ($\cdot$)\\Conv2d ($c_i$,64,(5,5),3,0)\\BatchNorm2d (64)\\ReLU\\Conv2d (64,128,(3,3),1,0)\\BatchNorm2d (128)\\ReLU\\Flatten\\FC (128*8*8,1024)\\BatchNorm1d (1024)\\ReLU\\FC (1024,10) \end{tabular} & \begin{tabular}[c]{@{}l@{}}$\textsf{Normalized}$ ($\cdot$)\\Conv2d ($c_i$,64,(5,5),3,0)\\BatchNorm2d (64)\\ReLU\\Conv2d (64,128,(3,3),1,0)\\BatchNorm2d (128)\\ReLU\\Flatten\\FC (128*8*8,1024)\\BatchNorm1d (1024)\\ReLU\\FC (1024,$n_\text{o}$) \end{tabular} & \begin{tabular}[c]{@{}l@{}}$\textsf{Normalized}$ ($\cdot$)\\Conv2d ($n_\text{i}$,128,(3,3),1,0)\\BatchNorm2d (128)\\ReLU\\Flatten\\FC (128*8*8,1024)\\BatchNorm1d (1024)\\ReLU\\FC (1024,10) \end{tabular} & \begin{tabular}[c]{@{}l@{}}$\textsf{Normalized}$ ($\cdot$)\\Conv2d ($n_\text{i}$,128,(3,3),1,0)\\BatchNorm2d (128)\\ReLU\\Flatten\\FC (128*8*8,1024)\\BatchNorm1d (1024)\\ReLU\\FC (1024,$n_\text{o}$) \end{tabular}  \\ 
				\midrule
				$\B{\bar{g}_{\Post}}$  & Identity function                                                                                                                                                                                                                                     & \begin{tabular}[c]{@{}l@{}}ReLU\\FC ($n_\text{o}$,10) \end{tabular}                                                                                                                                                                                             & Identity function                                                                                                                                                                                          & \begin{tabular}[c]{@{}l@{}}ReLU\\FC ($n_\text{o}$,10) \end{tabular}                                                                                                                                                   \\
				\bottomrule
			\end{tabular}
			%\end{sc}
		\end{small}
	\end{center}
	%\vskip -0.1in                           	
\end{table*}

For the classification and the autoencoder tasks, the proposed algorithm is evaluated for MNIST \cite{lecun2010mnist}, Fashion-MNIST \cite{Fmnist}, and Cifar-10 \cite{Krizhevsky09learningmultiple} datasets by using their standard training sets and testing sets. Furthermore, for the autoencoder task, we evaluate the method for CelebA dataset~\cite{liu2015faceattributes}. The only used preprocessings on images are Random Crop and Random Horizontal Flip on Cifar-10 training dataset for the classification task, and padding MNIST and Fashion-MNIST images on all sides with zeros of length $4$ to fit in a network with input size $32 \times 32$. We set the training batch size equal to 128.

We employ Convolutional Neural Networks (CNNs) in $\bar{g}_{\Pre}$, $f_j$, and $\bar{g}_{\Post}$. We initialize the network parameters by Kaiming initialization \cite{KHeI}. 
For each value of the standard deviation of the noise, we continue the learning process until the loss converges, i.e., when the difference in consecutive losses diminishes. We also use Adam optimizer~\cite{AdamOpt} and decrease the learning rate from $10^{-3}$ to $2 \times 10^{-5}$ gradually during the training.
For $N=2$ and $T=1$, we choose
\begin{align*}
\M{W}_{1 \times 2}=
\begin{bmatrix}
1 & -1
\end{bmatrix}.
\end{align*}
For $N=3$ and $T=2$, we choose
\begin{align*}
\M{W}_{2 \times 3}=
\begin{bmatrix}
0 & \sqrt{\frac{3}{4}} & -\sqrt{\frac{3}{4}}\\
1 & -\frac{1}{2} & -\frac{1}{2}
\end{bmatrix}.
\end{align*}
For $N=4$ and $T=3$, we choose
\begin{align*}
\M{W}_{3 \times 4}=
\begin{bmatrix}
0 & \sqrt{\frac{8}{9}} & -\sqrt{\frac{2}{9}} & -\sqrt{\frac{2}{9}}\\
0 & 0 & \sqrt{\frac{2}{3}} & -\sqrt{\frac{2}{3}} \\
1 & -\frac{1}{3} & -\frac{1}{3} & -\frac{1}{3}
\end{bmatrix}.
\end{align*}
For $N=5$ and $T=2$, we choose
\begin{align*}
\M{W}_{2 \times 5}=
\begin{bmatrix}
\sin 0\alpha & \sin 1\alpha & \sin 2\alpha & \sin 3\alpha & \sin 4\alpha\\
\cos 0\alpha & \cos 1\alpha & \cos 2\alpha & \cos 3\alpha & \cos 4\alpha
\end{bmatrix},
\end{align*}
where $\alpha = \frac{2\pi}{5}$.

\subsection{The Classification Task}\label{subCla}

\begin{figure}[!t]%icv:[ht]
	%\vskip 0.2in
	\centering
	\includegraphics[width=\columnwidth,height=6.5cm,keepaspectratio]{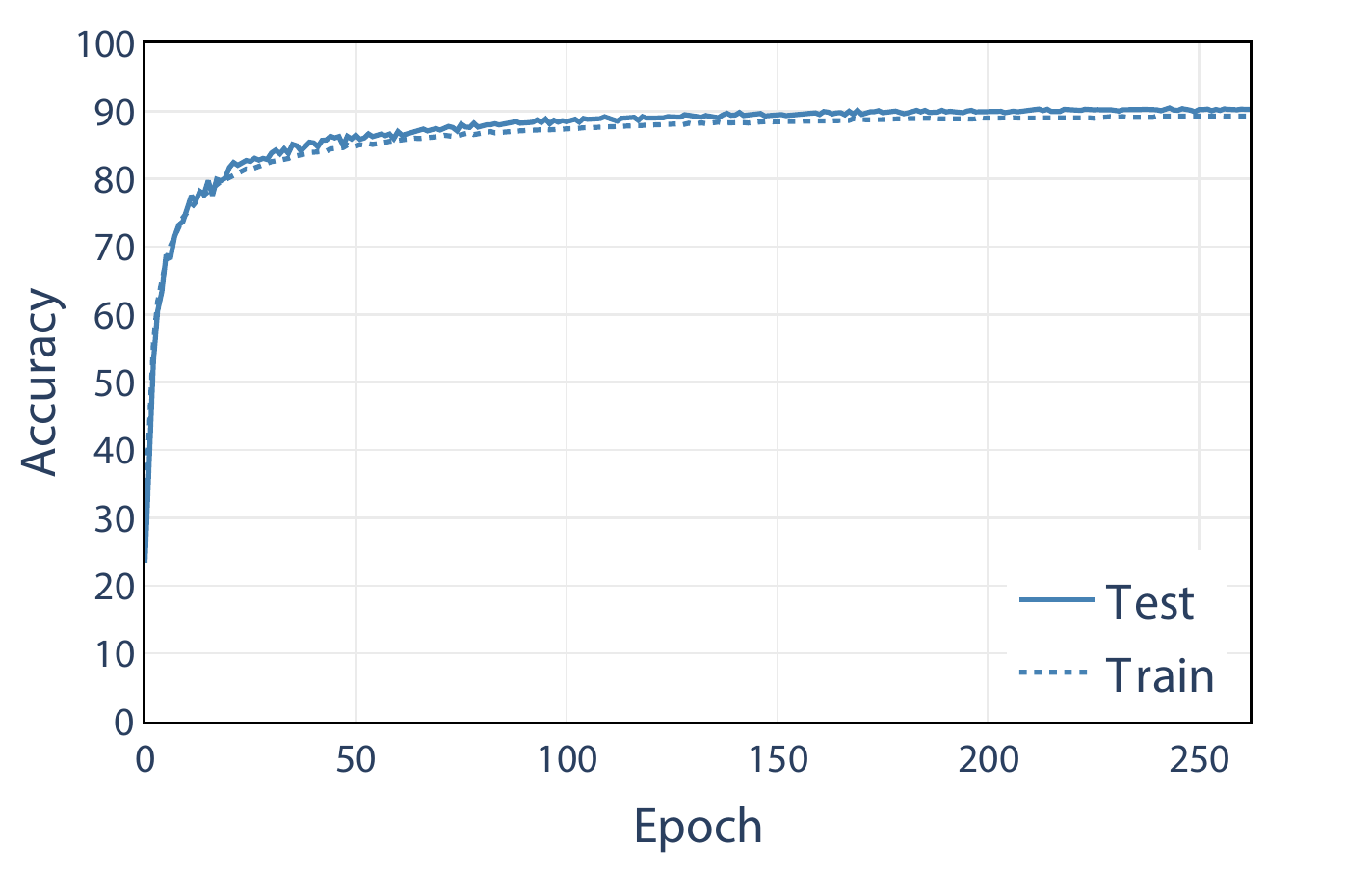}
	\caption{Results on the accuracy versus the epoch number, for noise level $\sigma_{\bar{Z}}=70$, $N=2$, $T=1$, and $\net( \mathsf{Iden.} \rightarrow \mathsf{Iden.})$ model on MNIST dataset in Experiment 1.}
	\label{fig:accepoch}
	%\vskip -0.2in
\end{figure}

\begin{figure*}[!t]%icv:[ht]%[ht]%[btp]%[ht]
	%\vskip 0.2in
	\centering
	\begin{subfigure}{0.68\textwidth}
		\centering
		\includegraphics[width=\linewidth,height=6.5cm,keepaspectratio]{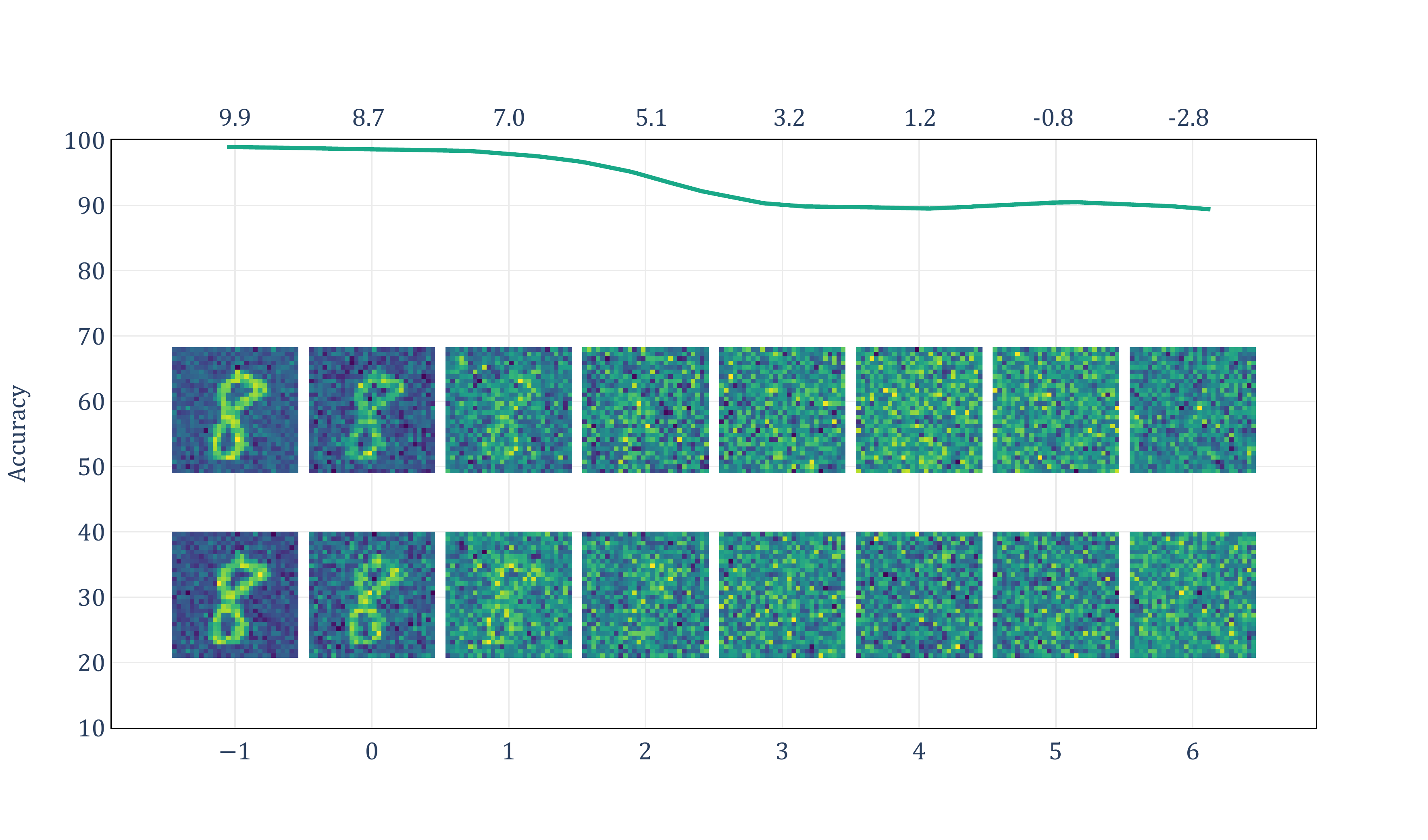}
		%\vskip -0.1in
		\caption{Accuracy of MNIST testing dataset versus $\log \sigma_{\bar{Z}}$ and $\log \varepsilon$ for $\net( \mathsf{Iden.} \rightarrow \mathsf{Iden.})$ model, $N=2$, and $T=1$. Also, it shows the inputs of both servers for a testing sample with label $"8"$ for $\log \sigma_{\bar{Z}}=-1,0,\ldots,6$.}
		\label{fig:Vir_im}
	\end{subfigure}%
	\hfill
	\begin{subfigure}{0.3\textwidth}
		\centering
		\vskip 0.37in
		\includegraphics[width=\linewidth,height=4.5cm,keepaspectratio]{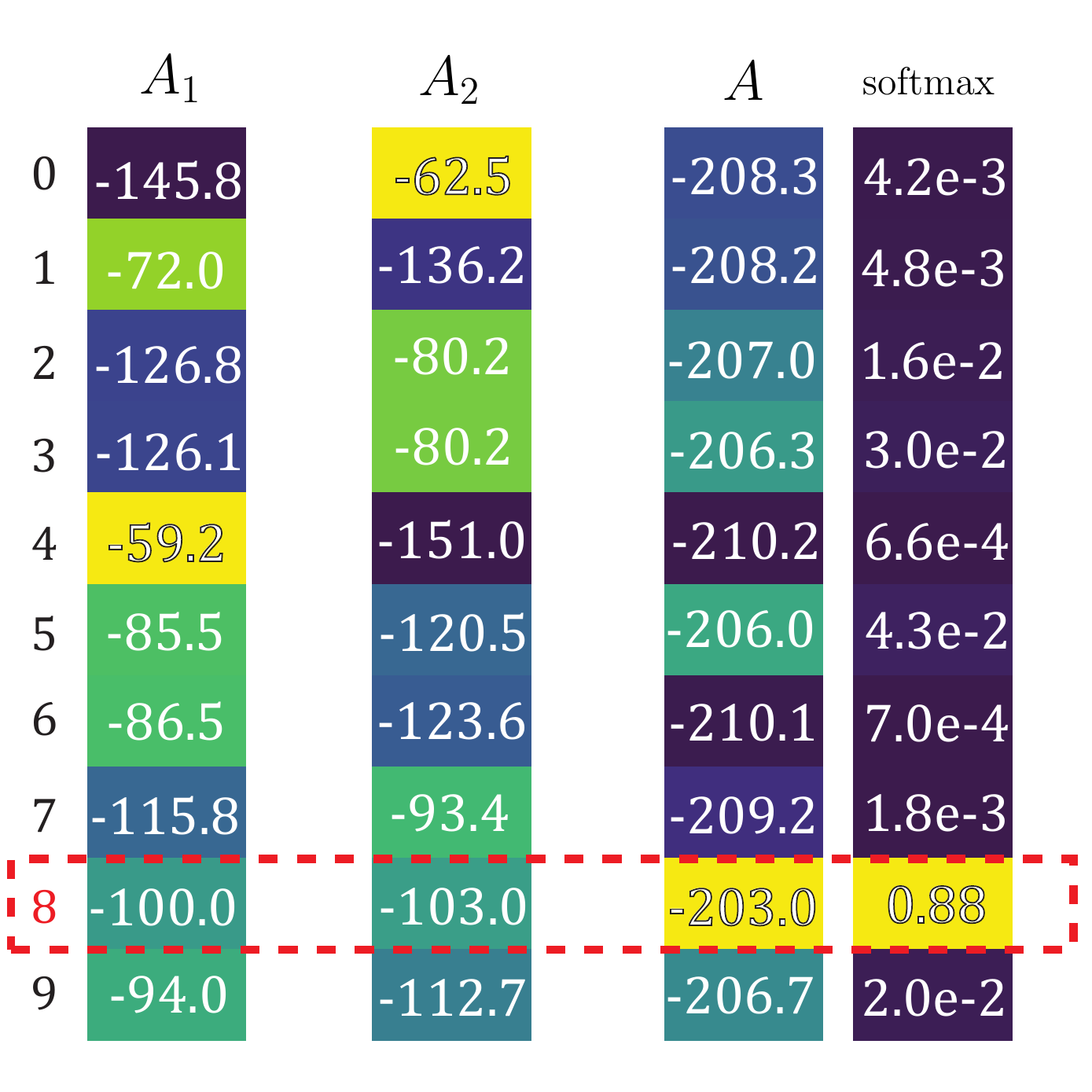}
		\vskip 0.37in
		\caption{The servers output and summation of them with and without applying softmax function for a sample with label $"8"$ for $\sigma_{\bar{Z}}=70$.}
		\label{fig:Vir_out}
	\end{subfigure}%
	%\vskip -0.1in
	\caption{The results of Experiment 1.}\label{fig:Vir}
	\vskip -0.1in
\end{figure*}

\begin{figure}[!t]%icv:[ht]
	%\vskip 0.2in
	\centering
	\includegraphics[width=\columnwidth,height=6cm,keepaspectratio]{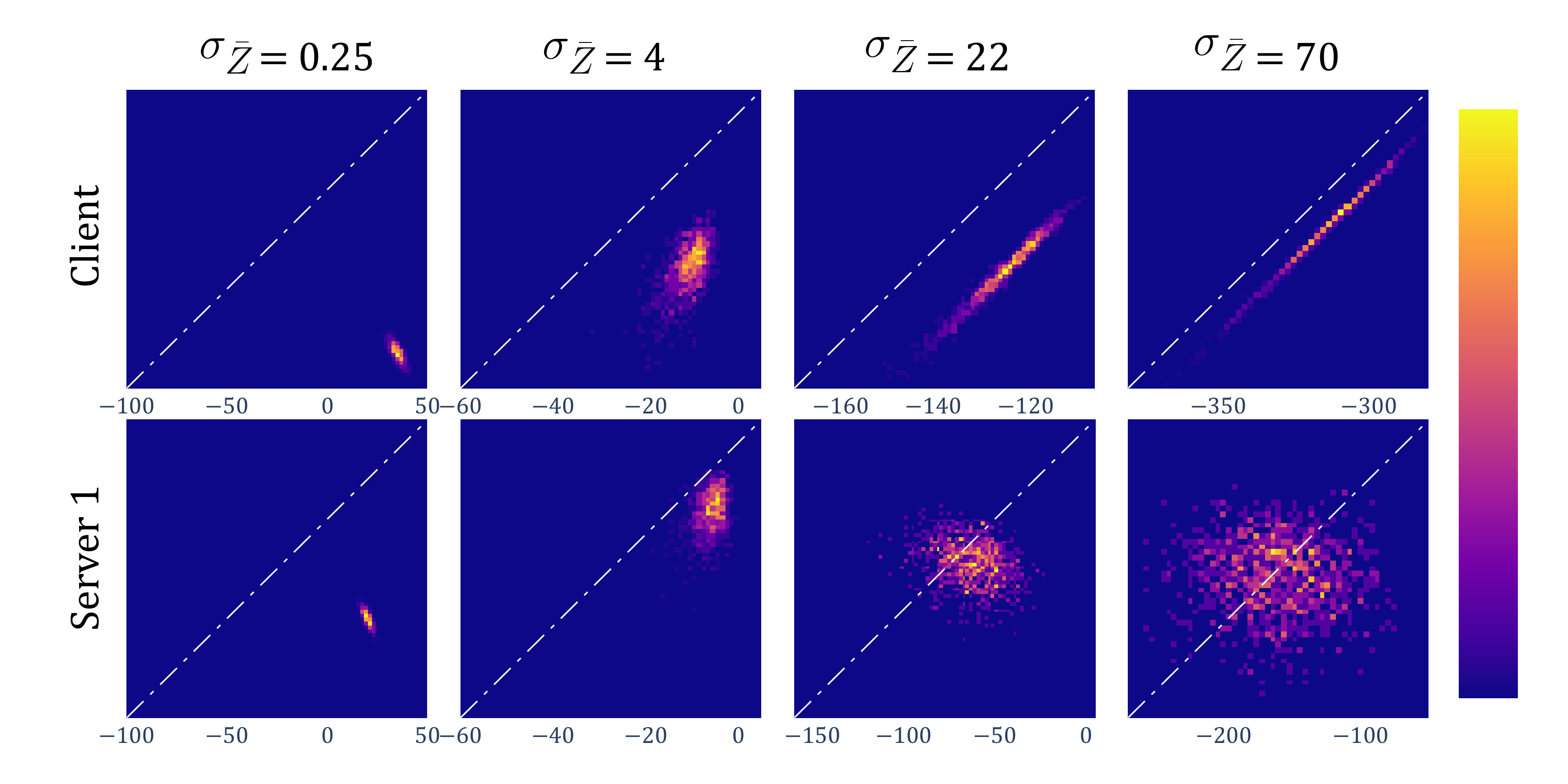}
	\caption{The joint distribution of $(A[6],A[9])$ in the \textbf{first row} and $(A_1[6],A_1[9])$ in the \textbf{second row} for a sample with label $"6"$ for various the standard deviation of the noise (i.e., $\sigma_{\bar{Z}}$) in Expriment 1.}
	\label{fig:Dist}
	%\vskip -0.2in
\end{figure}

\begin{figure*}[!t]%icv:[ht]
	%\vskip 0.2in
	\centering
	\begin{subfigure}{.333\textwidth}%
		\centering
		\includegraphics[width=\linewidth,height=7cm,keepaspectratio]{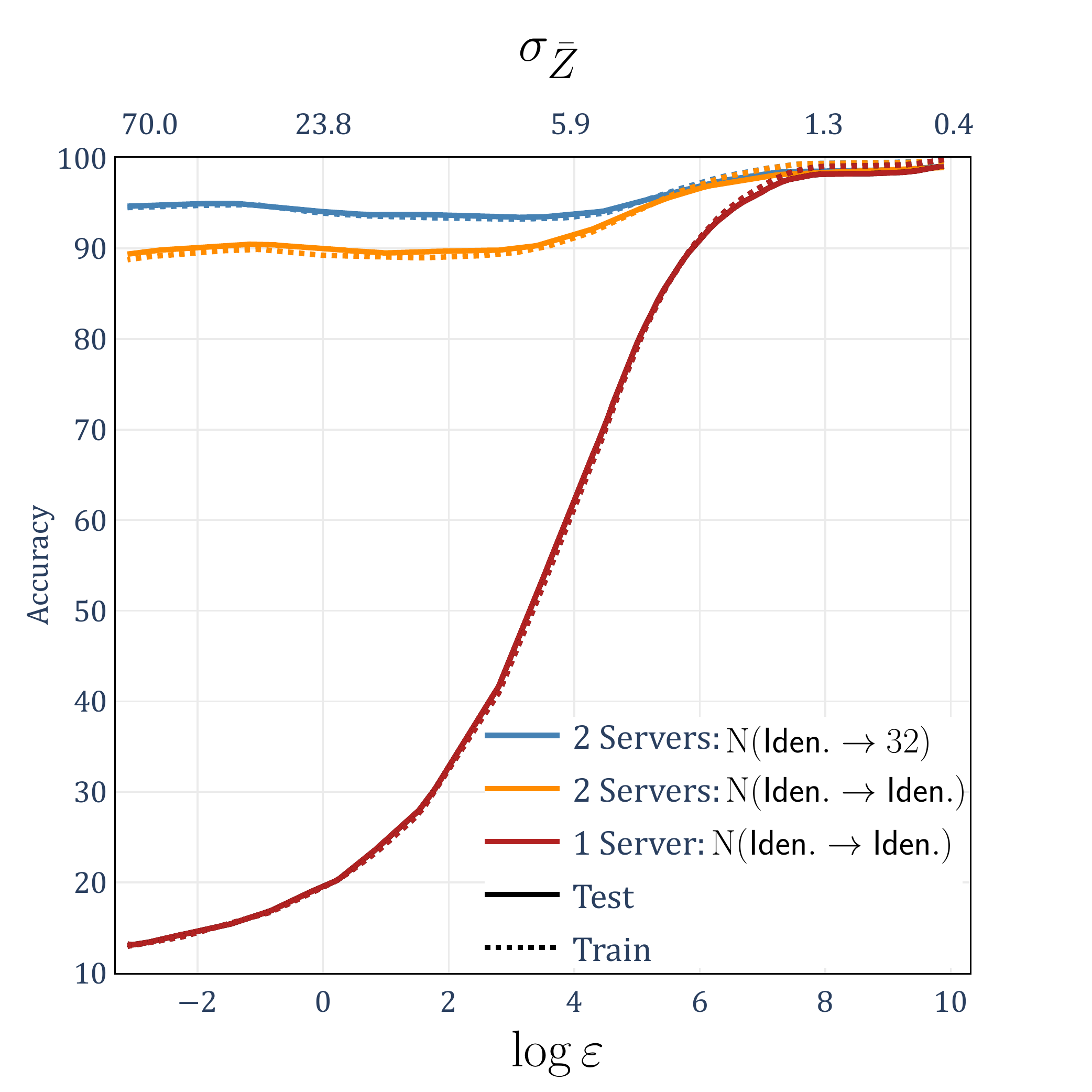}
		\caption{MNIST}
		\label{fig:secMNIST}
	\end{subfigure}%
	\hfill
	\begin{subfigure}{.333\textwidth}
		\centering
		\includegraphics[width=\linewidth,height=7cm,keepaspectratio]{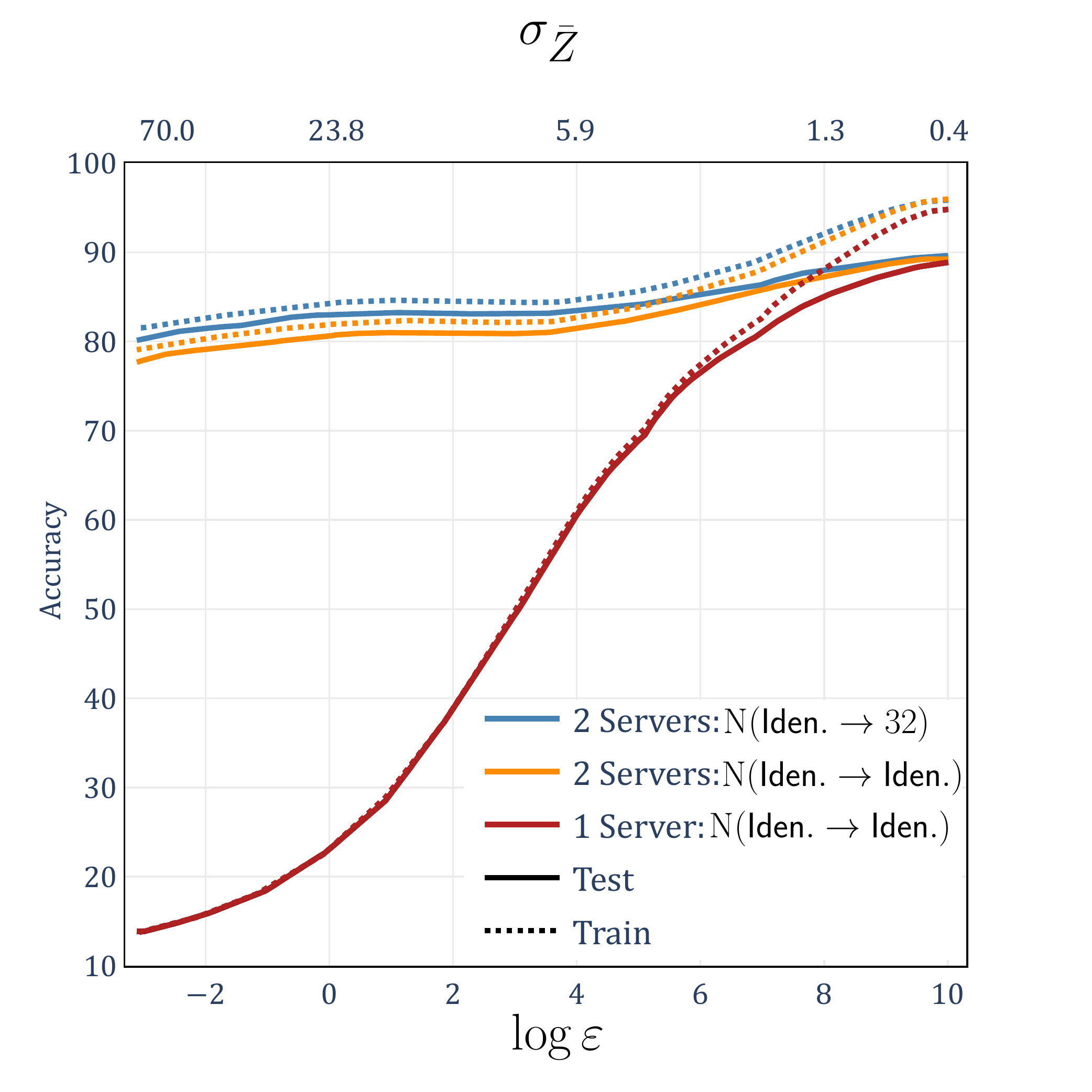}
		\caption{Fashion-MNIST}
		\label{fig:secFashionMNIST}
	\end{subfigure}%
	\hfill
	\begin{subfigure}{.333\textwidth}
		\centering
		\includegraphics[width=\linewidth,height=7cm,keepaspectratio]{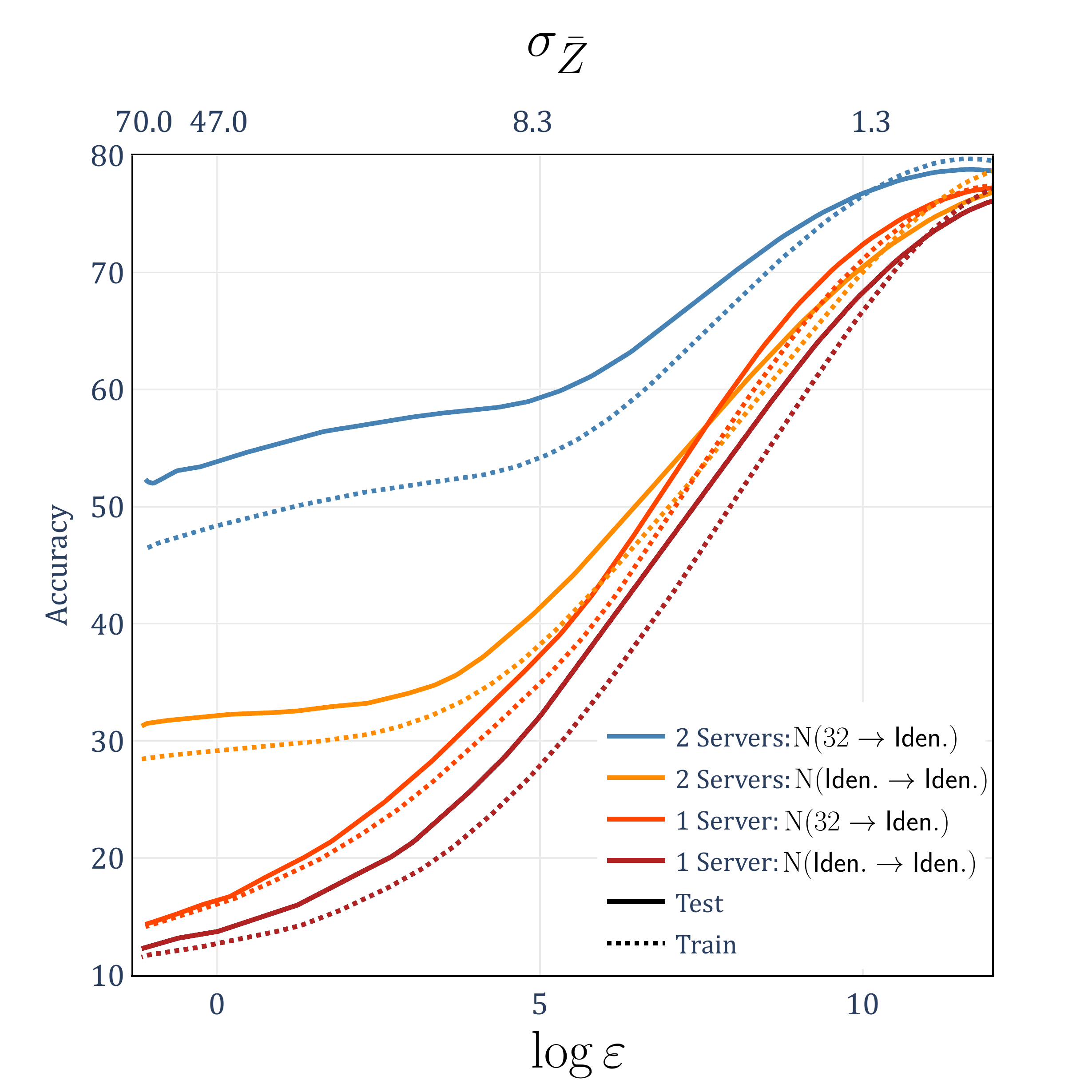}
		\caption{Cifar-10}
		\label{fig:secCifar}
	\end{subfigure}%
	\hfill
	\begin{subfigure}{.333\textwidth}
		\centering
		\includegraphics[width=\linewidth,height=7cm,keepaspectratio]{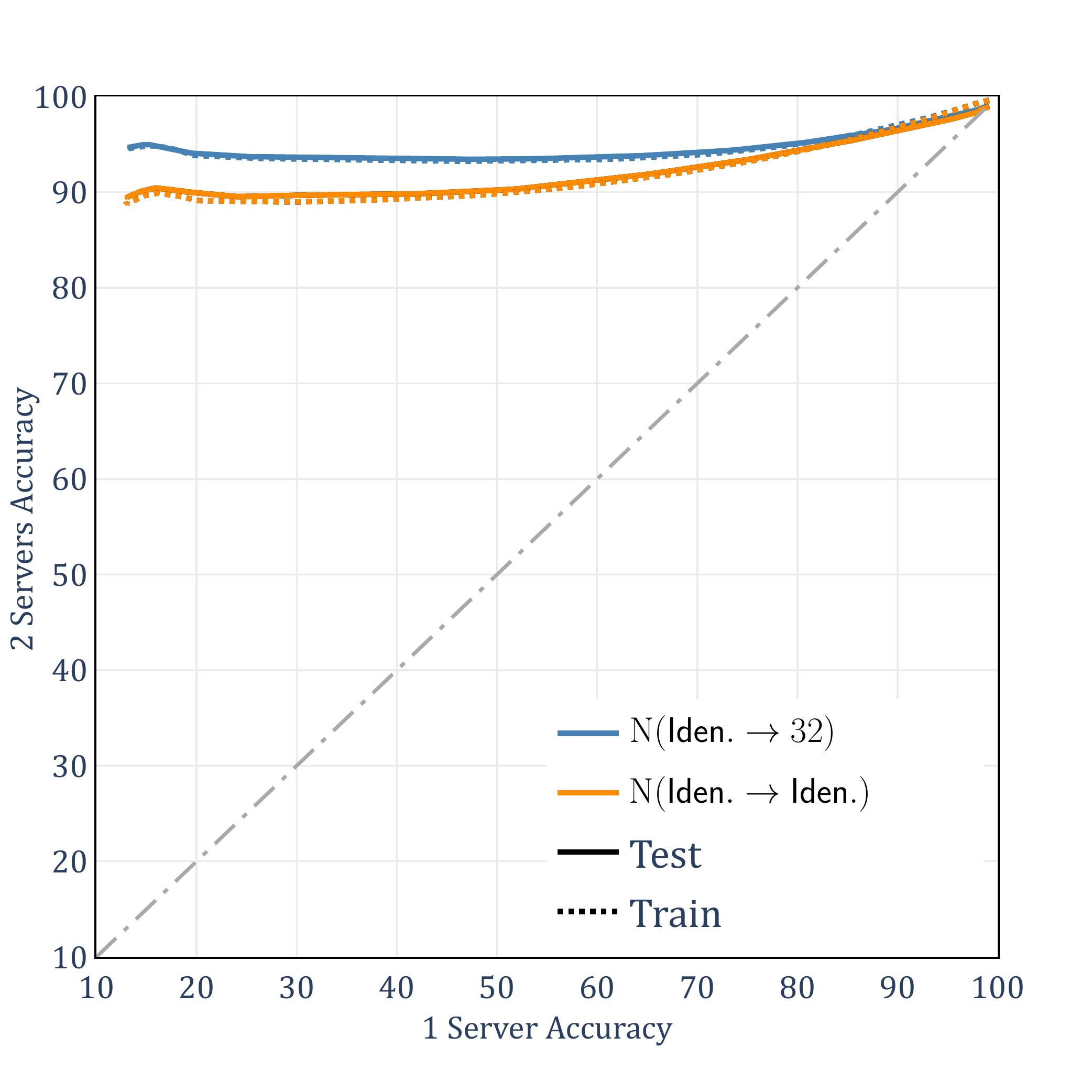}
		\caption{MNIST}
		\label{fig:rocMNIST}
	\end{subfigure}%
	\hfill
	\begin{subfigure}{.333\textwidth}
		\centering
		\includegraphics[width=\linewidth,height=7cm,keepaspectratio]{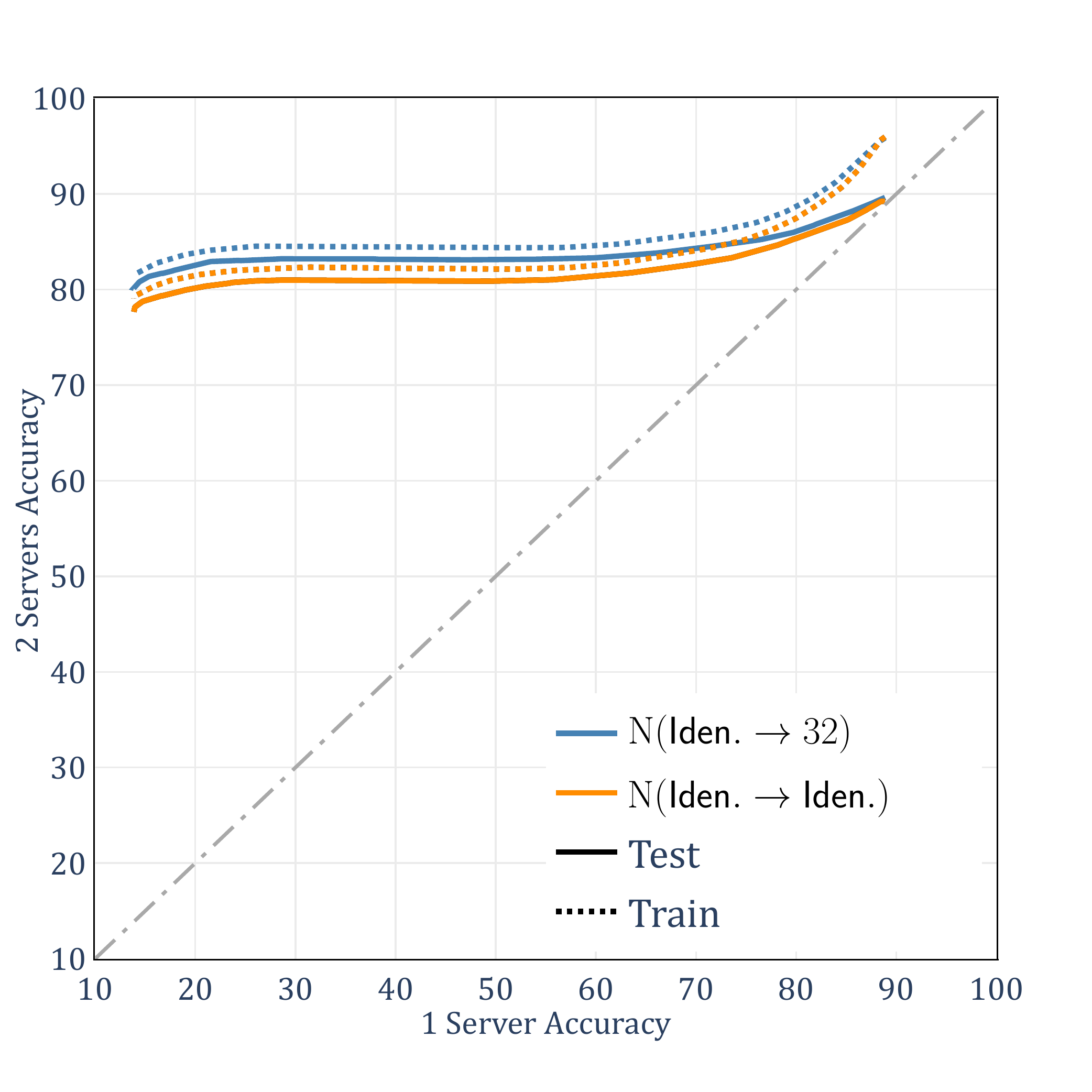}
		\caption{Fashion-MNIST}
		\label{fig:rocFashionMNIST}
	\end{subfigure}%
	\hfill
	\begin{subfigure}{.333\textwidth}
		\centering
		\includegraphics[width=\linewidth,height=7cm,keepaspectratio]{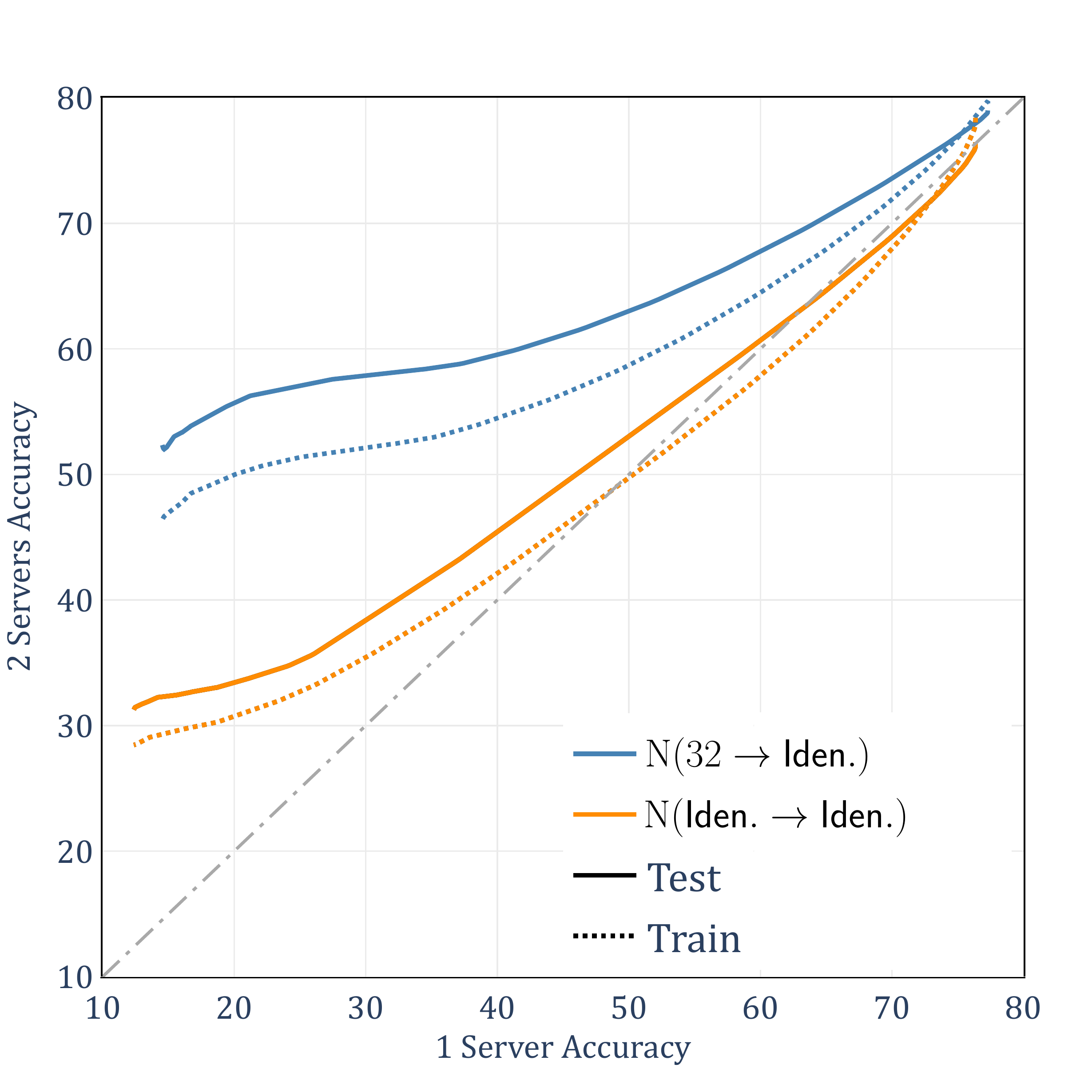}
		\caption{Cifar-10}
		\label{fig:rocCifar}
	\end{subfigure}%
	\caption{The privacy and accuracy curves (The results of Experiment 2)}\label{fig:secroc}
	%\vskip -0.2in
\end{figure*}

In this setup, $f_j$ is a neural network with several convolutional layers and two fully connected layers with the Rectified Linear Unit (ReLU) activation function (see Table~\ref{tab:tabStruc} for details). To limit the computation cost of the pre- and post-processings at the client, we use at most one convolutional layer in $\bar{g}_{\Pre}$ and at most one fully connected layer in $\bar{g}_{\Post}$. In particular, $\bar{g}_{\Pre}$ is either an identity function, denoted by $\mathsf{Iden.}$, or a convolutional layer with $n_\text{i} \in \mathbb{N}$ output channels. In addition, $\bar{g}_{\Post}$ can be a fully connected layer, with a vector of length $n_\text{o} \in \mathbb{N}$ as the input, and generating $10$ outputs, representing $10$ different classes. The input vector of length $n_\text{o}$ is formed by adding the $N$ vectors of length $n_\text{o}$, received the from servers. In addition, we also consider a very simple case for $\bar{g}_{\Post}$, where $n_\text{o} = 10$ and at the client side, we simply add up the vectors of length $n_\text{o}$, received from the servers. In other words, in this case $\bar{g}_{\Post}$ is equal to the identity function. We represent the structure of $\bar{g}_{\Pre}$ and $\bar{g}_{\Post}$ by $\B{\net(n_\text{i}\rightarrow n_\text{o})}$, where the number of the output channels at $\bar{g}_{\Pre}$ is equal to $n_\text{i}$ (with the exception that $n_\text{i}=\mathsf{Iden.}$ means $\bar{g}_{\Pre}$ is the identity function, i.e., $\bar{g}_{\Pre}(\B{X}) = \B{X}$), and the number of the output neurons at $f_j$ is equal to $n_\text{o}$ (with the exception that $n_\text{o}=\mathsf{Iden.}$ means  $\bar{g}_{\Post}(A) = A$). In Table~\ref{tab:tabStruc}, $c_i$ denotes the number of the input image channels. Since the variance of the input queries can be high (see equation~\eqref{def:G}) we use $\Norma(.)$ function at the first stage of $f_j$. We use the cross-entropy loss function between $\B{Y}$ and $\mathsf{softmax}(\B{\hat{Y}})$ for $\Loss\{\B{\hat{Y}},\B{Y}\}$.

\subsubsection{\textbf{Experiment 1}}

In this experiment, we focus on MNIST dataset, and consider a system with $N=2$ servers, no colluding ($T=1$). In addition, we use identity mapping for both $\bar{g}_{\Pre}$ and $\bar{g}_{\Post}$. This means the number of the output neurons at the servers is $10$ and we simply add the received answers of the servers to find the label of the client data. This case is denoted by $\net( \mathsf{Iden.} \rightarrow \mathsf{Iden.})$. Fig.~\ref{fig:accepoch} reports the accuracy for the testing and training dataset versus the training epoch number for $\sigma_{\bar{Z}}=70$, representing a very strong noise. This figure shows that the proposed approach allows the model of the servers to train in presence of the strong but correlated noises, such that the client can mitigate the contribution of the noises by combining the outputs of the servers.

Fig.~\ref{fig:Vir_im} reports the accuracy for the testing dataset at the end of the training phase versus $\sigma_{\bar{Z}}$ and $\varepsilon$ for different $\sigma_{\bar{Z}}$ ranging from $0$ to $70$. It shows that using Corella, the client achieves $90\%$ accuracy while the privacy leakage is less than $\varepsilon = 0.12$, thanks to the strong noise with $\sigma_{\bar{Z}} = 70$. 
Also, this figure visualizes the inputs and the outputs of the servers for an MNIST sample in this experiment. As you can see, when $\log \sigma_{\bar{Z}} = 6$, the input of each server is severely noisy, and we cannot identify the label of the input from it. Still, the client can detect the label with high accuracy.
Fig.~\ref{fig:Vir_out} shows the answer of each server (which is a vector of length $10$), where the correct label is $"8"$ and $\sigma_{\bar{Z}}=70$. As you can see, in the answer of an individual server ($A_1$ and $A_2$), the correct label (here $"8"$) does not rank among the labels with highest values. However, when we add the values correspondingly, $"8"$ has the highest value in the result. In particular, after applying softmax, the density on the correct label is significantly higher than the rest. This observation intuitively confirms the privacy and accuracy of the proposed Corella method.

A similar effect is shown in Fig.~\ref{fig:Dist}. In this figure, we use a test sample from MNIST dataset with label $"6"$ as the input. We want to visualize how each server is confused about the correct label (here $"6"$), with an incorrect one (say $"9"$). Each plot in the second row of Fig.~\ref{fig:Dist} is a 2D-plot histogram, representing the joint distribution of two neurons of the output of server one, i.e., $A_1[6]$ and $A_1[9]$ ($A_1[6]$ on the x-axis and $A_1[9]$ on the y-axis). We have this figure for different values of $\sigma_{\bar{Z}}$. If the point $(A_1[6],  A_1[9])$ is above the line $y=x$, i.e., $A_1[9] > A_1[6]$, it means that server one incorrectly prefers label $"9"$ to label $"6"$. 
In the first row in Fig.~\ref{fig:Dist}, we have the same plots for $A[9]$ versus $A[6]$, where $A=A_1+A_2$. As we can see, even for large noise (i.e., $\sigma_{\bar{Z}}=70$), server one chooses $"6"$ or $"9"$ almost equiprobably, while the client almost always chooses the label correctly. This shows that, in our proposed method, simultaneous training of the two networks $f_1$ and $f_2$ on the correlated queries allows the system to be trained such that the distribution of the noise at the output of the system does not confuse the client about the correct label (see the first row in Fig.~\ref{fig:Dist}).

\subsubsection{\textbf{Experiment 2}}

In this part, we evaluate the performance of the proposed method for $N=2$ and $T=1$, for three different datasets, including MNIST, Fashion-MNIST, and CIFAR-10, and for various levels of noise. In addition, we compare the performance of the method, with the case where we have only one server and we add noise to the input to protect privacy of the data. For the system with one server, we train the model, in the presence of noise. For each dataset, we plot the accuracy of the proposed method for two models and $\sigma_{\bar{Z}}$ = $0$ to $70$. Figures~\ref{fig:secMNIST}, \ref{fig:secFashionMNIST}, and \ref{fig:secCifar} demonstrate both the accuracy of the proposed method and the accuracy of the system with one server, versus $\log \varepsilon$ and $\sigma_{\bar{Z}}$. Figures~\ref{fig:rocMNIST}, \ref{fig:rocFashionMNIST}, and \ref{fig:rocCifar} compare the accuracy of the proposed method with the case with one server. Figures~\ref{fig:secMNIST}, \ref{fig:secFashionMNIST}, and \ref{fig:secCifar} show that unlike the systems with one server, Corella achieves a good accuracy for various datasets, even for high level of noises. For example, in Fig.~\ref{fig:secMNIST}, the client with low post-processing in $N( \mathsf{Iden.} \rightarrow32)$ achieves $95\%$ accuracy for $\sigma_{\bar{Z}} = 70$, while with a single server and adding noise with the same variance, we can achieve $13\%$ accuracy (see Fig.~\ref{fig:rocMNIST}). This comparison emphasizes on the fact that having more than one server with correlated noise world allow the system to mitigate the contribution of the noise, even when the noise is strong. However with one server, even with training, the system cannot eliminate the noise. Thus, the ability of noise mitigation in Corella follows not only from training in the presence of noise but also the fact that we have more than one server and the noises are correlated. In general, although the accuracy of Corella decreases with increasing the variance of the noise, it still converges to a reasonable value.

\subsubsection{\textbf{Experiment 3}}

\begin{table*}[!t]%icv:%[t]
	\caption{Test accuracy in the classification task, for various models and different tuple $(N,T)$. The test accuracy for MNIST, Fashion-MNIST, and Cifar-10 is evaluated for $\log \varepsilon =$ 0, 0, and 1.5, respectively (The results of Experiment 3).}
	\label{tab:tabResults}
	%\vskip 0.1in
	\begin{center}
		\begin{small}
			%\begin{sc}
			\centering
			\begin{tabular}{lllllllll} 
				\toprule
				\multicolumn{1}{c}{\multirow{2}{*}{Dataset}} & \multicolumn{1}{c}{\multirow{2}{*}{Model}}       & \multicolumn{1}{c}{\multirow{2}{*}{\begin{tabular}[c]{@{}c@{}}$\frac {s}{\text{Image Size}}$\\ \end{tabular}}} & \multicolumn{1}{c}{\multirow{2}{*}{$\frac {\Cc (g)}{\Cc (f_j)}$ }} & \multicolumn{1}{c}{\multirow{2}{*}{$\frac {\Cs (g)}{\Cs (f_j)}$ }} & \multicolumn{4}{c}{ Accuracy for $(N,T)$ }                                                                                                 \\ 
				\cmidrule(lr){6-9}
				\multicolumn{1}{c}{}                         & \multicolumn{1}{c}{}                             & \multicolumn{1}{c}{}                                                                                           & \multicolumn{1}{c}{}                                               & \multicolumn{1}{c}{}                                               & \multicolumn{1}{c}{$(2,1)$ } & \multicolumn{1}{c}{$(3,2)$ } & \multicolumn{1}{c}{$(4,3)$ } & \multicolumn{1}{c}{$(5,2)$ }                  \\ 
				\toprule
				\multirow{5}{*}{MNIST}                       & $\net(\mathsf{Iden.}\rightarrow\mathsf{Iden.})$  & 1                                                                                                              & 0                                                                  & 0                                                                  & 90.72                        & 90.52                        & 90.23                        & 91.02                                         \\ 
				\cdashline{2-9}[1pt/1pt]
				& $\net(\mathsf{Iden.}\rightarrow32)$              & 1                                                                                                              & 3.1e-5                                                             & 5.1e-5                                                             & 95.16                        & 95.10                        & 94.87                        & 95.78                                         \\
				& $\net(\mathsf{Iden.}\rightarrow64)$              & 1                                                                                                              & 6.3e-5                                                             & 9.9e-5                                                             & 96.40                        & 95.57                        & 94.82                        & 96.73                                         \\ 
				\cdashline{2-9}[1pt/1pt]
				& $\net(1\rightarrow\mathsf{Iden.})$               & 1.0e-1                                                                                                         & 3.1e-4                                                             & 4.0e-6                                                             & 90.53                        & -                            & -                            & -                                             \\
				& $\net(2\rightarrow\mathsf{Iden.})$               & 2.1e-1                                                                                                         & 6.2e-4                                                             & 8.1e-6                                                             & 94.21                        & -                            & -                            & -                                             \\ 
				\midrule
				\multirow{8}{*}{Fashion-MNIST}               & $\net(\mathsf{Iden.}\rightarrow\mathsf{Iden.})$  & 1                                                                                                              & 0                                                                  & 0                                                                  & 81.00                        & 80.45                        & 80.13                        & 81.39                                         \\ 
				\cdashline{2-9}[1pt/1pt]
				& $\net(\mathsf{Iden.}\rightarrow32)$              & 1                                                                                                              & 3.1e-5                                                             & 5.1e-5                                                             & 83.32                        & 82.45                        & 82.04                        & 83.88                                         \\
				& $\net(\mathsf{Iden.}\rightarrow64)$              & 1                                                                                                              & 6.3e-5                                                             & 9.9e-5                                                             & 84.00                        & -                            & -                            & -                                             \\
				& $\net(\mathsf{Iden.}\rightarrow128)$             & 1                                                                                                              & 1.2e-4                                                             & 1.9e-4                                                             & 83.02                        & -                            & -                            & -                                             \\ 
				\cdashline{2-9}[1pt/1pt]
				& $\net(2\rightarrow\mathsf{Iden.})$               & 2.1e-1                                                                                                         & 6.2e-4                                                             & 8.1e-6                                                             & 81.69                        & -                            & -                            & -                                             \\
				& $\net(4\rightarrow\mathsf{Iden.})$               & 4.1e-1                                                                                                         & 1.2e-3                                                             & 1.6e-5                                                             & 83.35                        & -                            & -                            & -                                             \\
				& $\net(8\rightarrow\mathsf{Iden.})$               & 8.3e-1                                                                                                         & 2.4e-3                                                             & 3.2e-5                                                             & 81.51                        & -                            & -                            & -                                             \\ 
				\cdashline{2-9}[1pt/1pt]
				& $\net(4\rightarrow32)$                           & 4.1e-1                                                                                                         & 1.3e-3                                                             & 6.7e-5                                                             & 83.13                        & -                            & -                            & -                                             \\ 
				\midrule
				\multirow{9}{*}{Cifar-10}                    & $\net(\mathsf{Iden.}\rightarrow\mathsf{Iden.})$  & 1                                                                                                              & 0                                                                  & 0                                                                  & 35.70                        & 35.47                        & 35.12                        & 36.82                                         \\ 
				\cdashline{2-9}[1pt/1pt]
				& $\net(\mathsf{Iden.}\rightarrow32)$              & 1                                                                                                              & 2.3e-5                                                             & 3.9e-5                                                             & 38.30                        & -                            & -                            & -                                             \\
				& $\net(\mathsf{Iden.}\rightarrow64)$              & 1                                                                                                              & 4.7e-5                                                             & 7.6e-5                                                             & 42.47                        & -                            & -                            & -                                             \\
				& $\net(\mathsf{Iden.}\rightarrow128)$             & 1                                                                                                              & 9.3e-5                                                             & 1.5e-4                                                             & 42.28                        & -                            & -                            & -                                             \\ 
				\cdashline{2-9}[1pt/1pt]
				& $\net(8\rightarrow\mathsf{Iden.})$               & 2.6e-1                                                                                                         & 6.7e-3                                                             & 7.2e-5                                                             & 50.30                        & -                            & -                            & -                                             \\
				& $\net(16\rightarrow\mathsf{Iden.})$              & 5.2e-1                                                                                                         & 1.3e-2                                                             & 1.4e-4                                                             & 54.39                        & -                            & -                            & -                                             \\
				& $\net(32\rightarrow\mathsf{Iden.})$              & 1.0                                                                                                            & 2.2e-2                                                             & 2.9e-4                                                             & 58.13                        & -                            & -                            & -                                             \\
				& $\net(64\rightarrow\mathsf{Iden.})$              & 2.1                                                                                                            & 3.7e-2                                                             & 5.7e-4                                                             & 55.27                        & -                            & -                            & \begin{tabular}[c]{@{}l@{}}-\\ \end{tabular}  \\ 
				\cdashline{2-9}[1pt/1pt]
				& $\net(32\rightarrow128)$                         & 1.0                                                                                                            & 2.2e-2                                                             & 4.3e-4                                                             & 58.16                        & 58.02                        & 57.63                        & 60.39                                         \\
				\bottomrule
			\end{tabular}
			%\end{sc}
		\end{small}
	\end{center}
	%\vskip -0.1in                           	
\end{table*}

In this experiment, we discuss the client costs in terms of the computation, storage, and communication. In Table~\ref{tab:tabResults}, we evaluate the proposed algorithm for various models and values for $(N,T)$. We report the test accuracy for three datasets and the computation and storage costs of the client relative to the computation and storage costs of one of the servers. In this table, $\Cc(\cdot)$ and $\Cs(\cdot)$ denote the number of products (representing computational complexity) and the number of the parameters (representing storage complexity) in a model, respectively. In addition, we report the size of each query, denoted by $s$, relative to the size of the client data (the image size).

In this table, the difference in the cost of the client and each server (in terms of computational and storage complexity) is quite evident. On the other hand, the communication load between the client and each server is low, which is in the order of the size of $\B{X}$ (in the stage of sending query), and in the order of the size of $\B{Y}$ (in the stage of receiving answer). The low costs of computation and communication at the client side, and the guarantee of good accuracy make Corella framework suitable for practical use cases in mobile devices and the internet of things (IoT).

\subsection{The Autoencoder Task}\label{subAut}

\begin{table}[!t]%icv:%[t]
	\caption{The network structures of the encoder and the decoder. ConvT2d represents the transposed convolutional layer, and its parameters denote the number of the input channels, the number of the output channels, the kernel size, the stride, and the padding, respectively.
	}
	\label{tab:tab_encdec}
	%\vskip 0.1in
	\begin{center}
		\begin{small}
			%\begin{sc}
			\centering
			\begin{tabular}{cc} 
				\toprule
				Encoder($\cdot$)                                                                                                                                                                                                                                           & Decoder($\cdot$)                                                                                                                                                                                                                     \\ 
				\toprule
				\multicolumn{1}{l}{\begin{tabular}[c]{@{}l@{}}$\textsf{Normalized}$ ($\cdot$)\\Conv2d ($c_i$,12,(4,4),2,1)\\BatchNorm2d (12)\\ReLU\\Conv2d (12,24,(4,4),2,1)\\BatchNorm2d (24)\\ReLU\\Conv2d (24,$c_o$,(4,4),2,1)\\BatchNorm2d ($c_o$)\\ReLU \end{tabular}} & \multicolumn{1}{l}{\begin{tabular}[c]{@{}l@{}}$\textsf{Normalized}$ ($\cdot$)\\ConvT2d ($c_o$,24,(4,4),2,1)\\BatchNorm2d (24)\\ReLU\\ConvT2d (24,12,(4,4),2,1)\\BatchNorm2d (12)\\ReLU\\ConvT2d (12,$c_i$,(4,4),2,1) \end{tabular}}  \\
				\bottomrule
			\end{tabular}
			%\end{sc}
		\end{small}
	\end{center}
	%\vskip -0.1in                           	
\end{table}

In this experiment, we evaluate the proposed method for the autoencoder task. The network structures of Encoder($\cdot$) and Decoder($\cdot$) are presented in Table~\ref{tab:tab_encdec}. In this table, $c_i$ and $c_o$ denote the number of channels of the input image (i.e., the client data) and the latent, respectively. As detailed in Table~~\ref{tab:tab_encdec}, the encoder includes three convolutional layers. Similarly, the decoder is composed of three transposed convolutional layers. 
We choose the network structures of $\bar{g}_{\Pre}$, $f_j$, and $\bar{g}_{\Post}$ as follows: (i) in offloading the encoding stage (Scenario 1), $\bar{g}_{\Pre}(\B{X})=\B{X}$, $f_j(Q_j)=$ Encoder($Q_j$), and $\bar{g}_{\Post}(A)=$ Decoder($A$); (ii) in offloading the decoding stage (Scenario 2), $\bar{g}_{\Pre}(\B{X})=$ Encoder($\B{X}$), $f_j(Q_j)=$ Decoder($Q_j$), and $\bar{g}_{\Post}(A)=A$; (iii) in 
offloading both encoding and decoding stages (Scenario 3), $\bar{g}_{\Pre}(\B{X})=\B{X}$, $f_j(Q_j)=$ Decoder(Encoder($Q_j$)), and $\bar{g}_{\Post}(A)=A$.
We use the mean squared error (MSE) loss between $\mathsf{sigmoid}(\B{\hat{Y}})$ and $\B{Y}$ for $\Loss\{\B{\hat{Y}},\B{Y}\}$, where $\B{Y}$ is equal to image $\B{X}$, $\B{\hat{Y}}$ denotes the output of the model, and the range of the pixels of image $\B{X}$ is $[0,1]$.

\subsubsection{\textbf{Experiment 4}}

\begin{figure*}[!t]%[ht]%[ht]%[btp]%[ht]
	%\vskip 0.2in
	\centering
	\begin{subfigure}{0.33\textwidth}%\linewidth
		\centering
		\includegraphics[width=\linewidth,height=9.2cm,keepaspectratio]{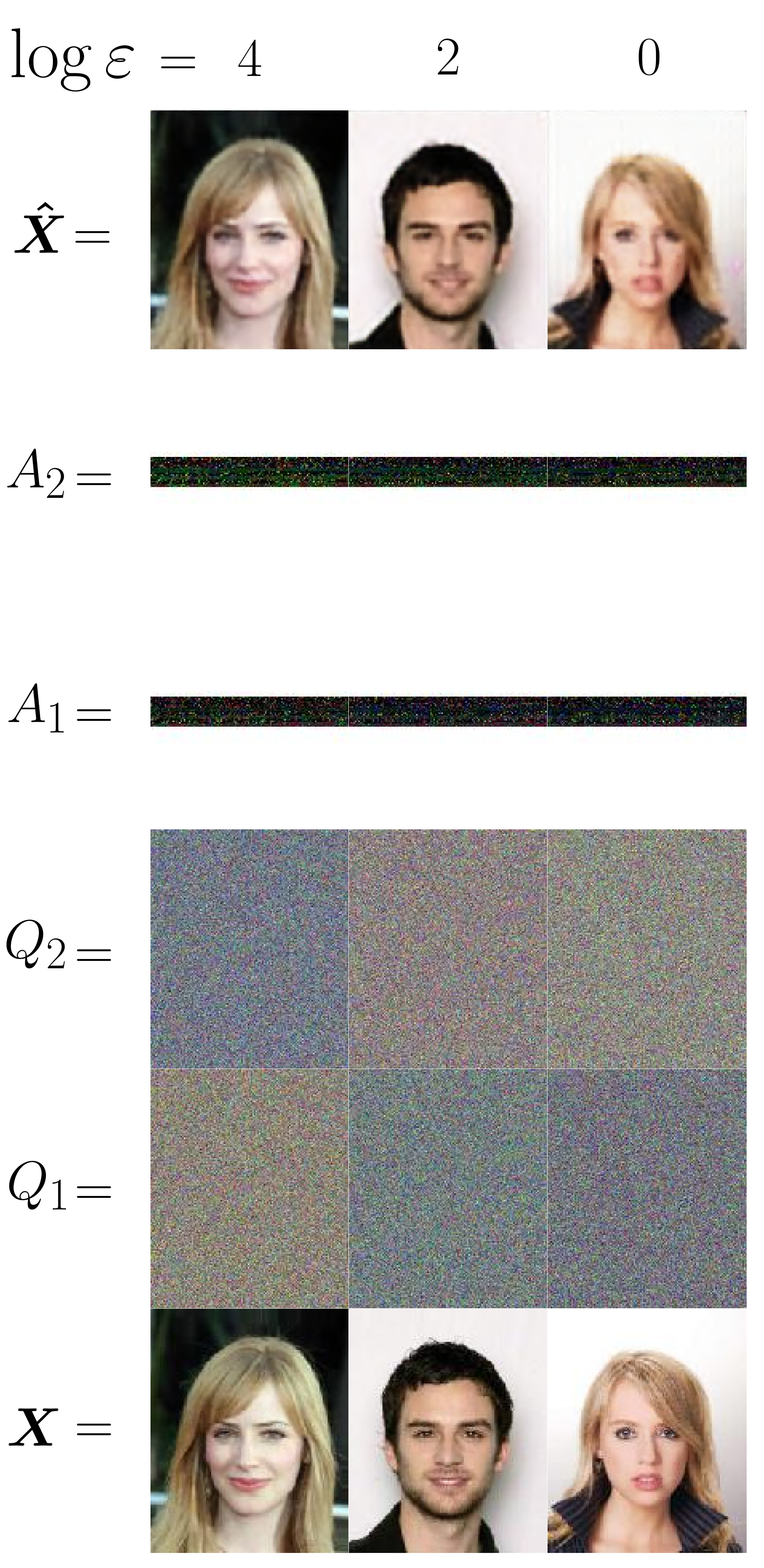}
		\caption{Scenario 1}
		\label{fig:CelebaImClient1}
	\end{subfigure}%
	\hfill
	\begin{subfigure}{0.3\textwidth}%\linewidth
		\centering
		\includegraphics[width=\linewidth,height=9.2cm,keepaspectratio]{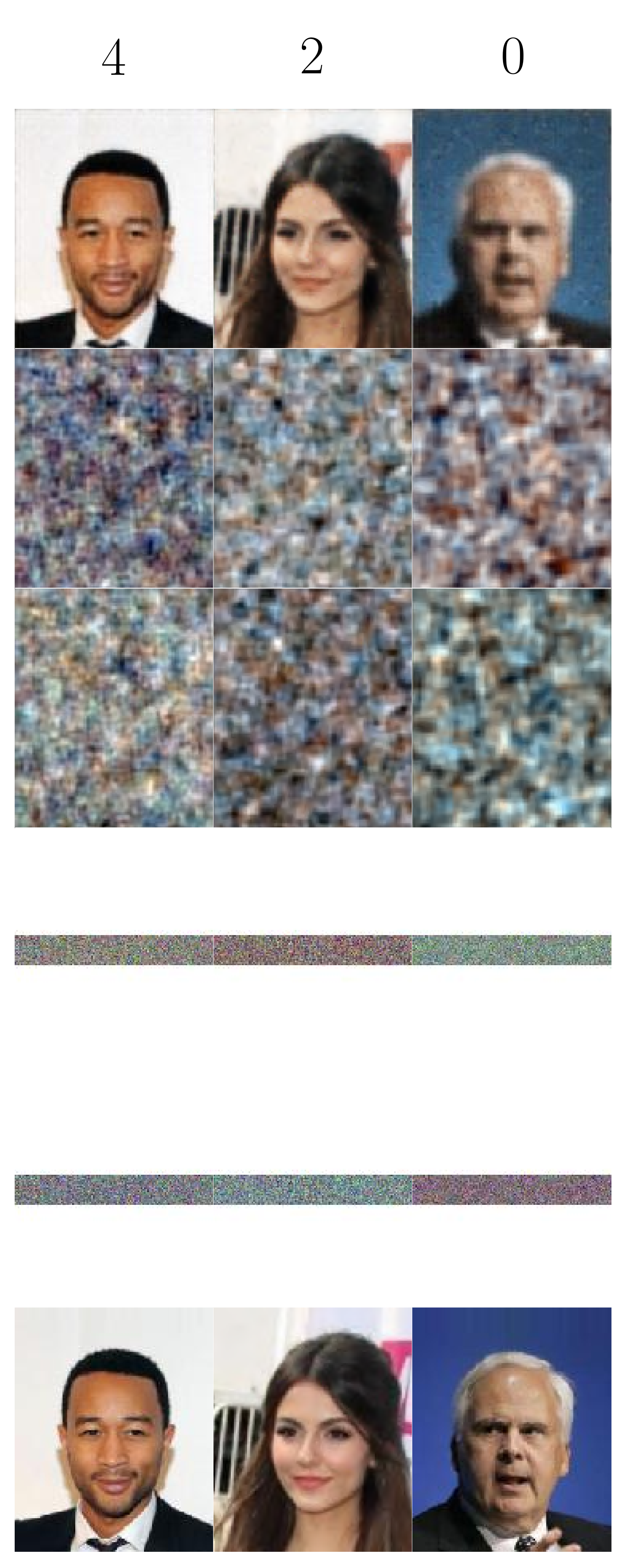}
		\caption{Scenario 2}
		\label{fig:CelebaImClient2}
	\end{subfigure}%
	\hfill
	\begin{subfigure}{0.3\textwidth}%\linewidth
		\centering
		\includegraphics[width=\linewidth,height=9.2cm,keepaspectratio]{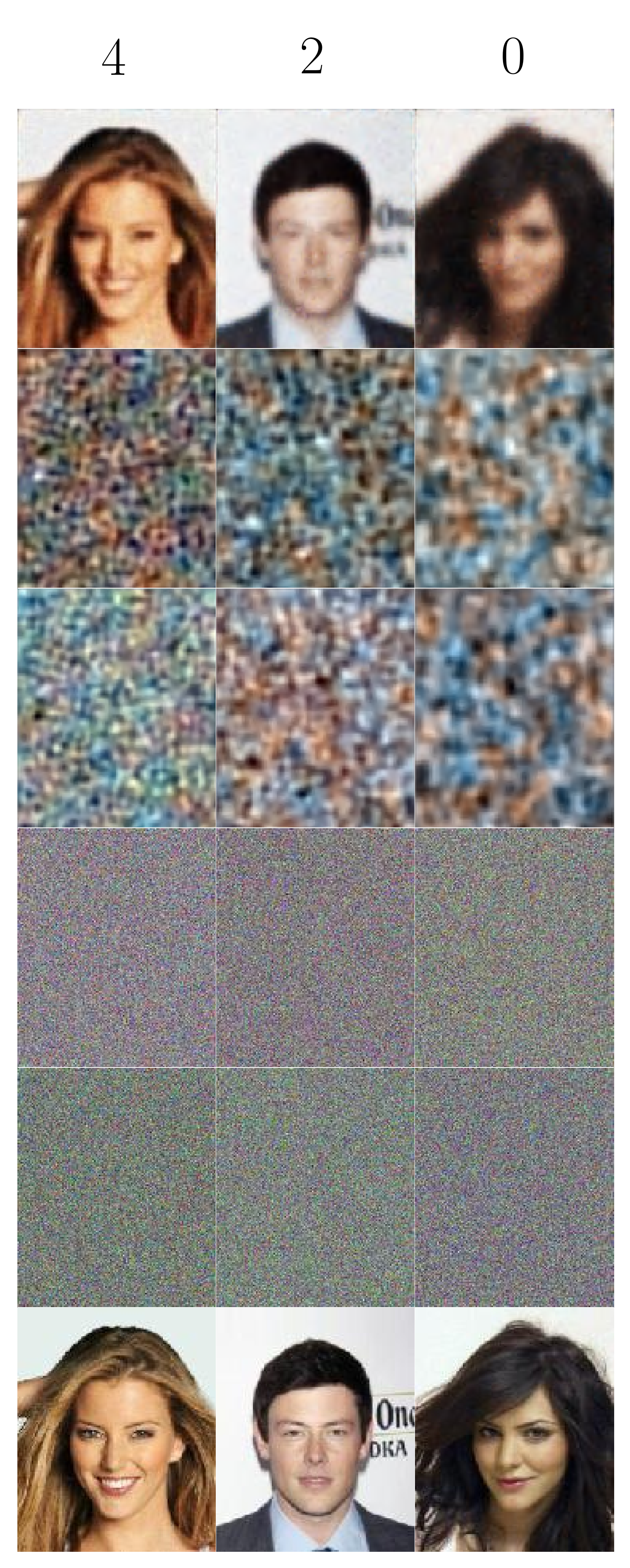}
		\caption{Scenario 3}
		\label{fig:CelebaImClient3}
	\end{subfigure}%
	%\vskip -0.1in
	\caption{The client image reconstruction. Visualization of the different the input images, the queries, the answers, and the reconstructed images in the three scenarios for compression ratio of $R_\Comp=8$, $(N,T)=(2,1)$, and $\log \varepsilon=0$, $2$, and $4$ (The results of Experiment 4).}\label{fig:CelebaImClient}
	%\vskip -0.2in
\end{figure*}

At first, we evaluate our proposed method for CelebA dataset for the compression ratio $R_\Comp=8$, $(N,T)=(2,1)$, and $\log \varepsilon = 0$, $2$, and $4$. When each layer of the encoder halves the height and width of its input (see Table~\ref{tab:tab_encdec}), the latent size is equal to $c_o \times \frac{h}{8} \times \frac{w}{8}$, where $h$ and $w$ represent the height and the width of the input image. Thus $R_\Comp=\frac{c_i \times h \times w}{c_o \times \frac{h}{8} \times \frac{w}{8}}=\frac{64 \times c_i}{c_o}$. As a result, for $R_\Comp=8$, we choose $c_o=24$, noting that CelebA images have 3 color channels (i.e., $c_i=3$).
Fig.~\ref{fig:CelebaImClient} visualizes the input image ($\B{X}$), the queries ($Q_1$ and $Q_2$), the answers ($A_1$ and $A_2$), and the reconstructed image ($\B{\hat{X}}=\mathsf{sigmoid}(\B{\hat{Y}})$), for different images from CelebA dataset. To plot the latent, we reshape it to an image with three channels. This figure shows that Corella method allows the client to offload privately the computational tasks of the encoder or the decoder, and then reconstruct the images, with high quality (shown in Fig.~\ref{fig:CelebaImClient1} and Fig.~\ref{fig:CelebaImClient2}), or offload both the computational tasks of the encoder and the decoder, and then reconstruct the images, with acceptable quality (shown in Fig.~\ref{fig:CelebaImClient3}).

\textbf{Remark~3:} 
For Scenario 3, one may suggest an alternative but naive method as follows.  We train two autoencoders separately, one to encode $\B{X}+\B{Z}$ and decode its latent to back $\B{X}+\B{Z}$, and the other to encode $\B{X}-\B{Z}$ and decode its latent to $\B{X}-\B{Z}$, where $\B{Z}$ is a noise with high variance. Then the client adds the outputs from the two decoders to reconstruct $\B{X}$. While this approach might seem reasonable, it doesn't work in practice. The reason is that $\B{X}+\B{Z}$ has a very high entropy, and thus cannot be compressed to a short latent and then be recovered accurately. The same statement holds for $\B{X}-\B{Z}$. The reason that Corella overcomes this problem is that in Corella, both servers are trained simultaneously such that the final result at the client is accurate. However, the input and the output of the autoencoder in each server are not similar at all. Fig.~\ref{fig:CelebaImClient3} demonstrates this observation clearly.

\begin{figure*}[!t]%[ht]%[ht]%[btp]%[ht]
	%\vskip -0.1in
	\centering
	\begin{subfigure}{0.36\textwidth}%\linewidth
		\centering
		\includegraphics[width=\linewidth,height=5.5cm,keepaspectratio]{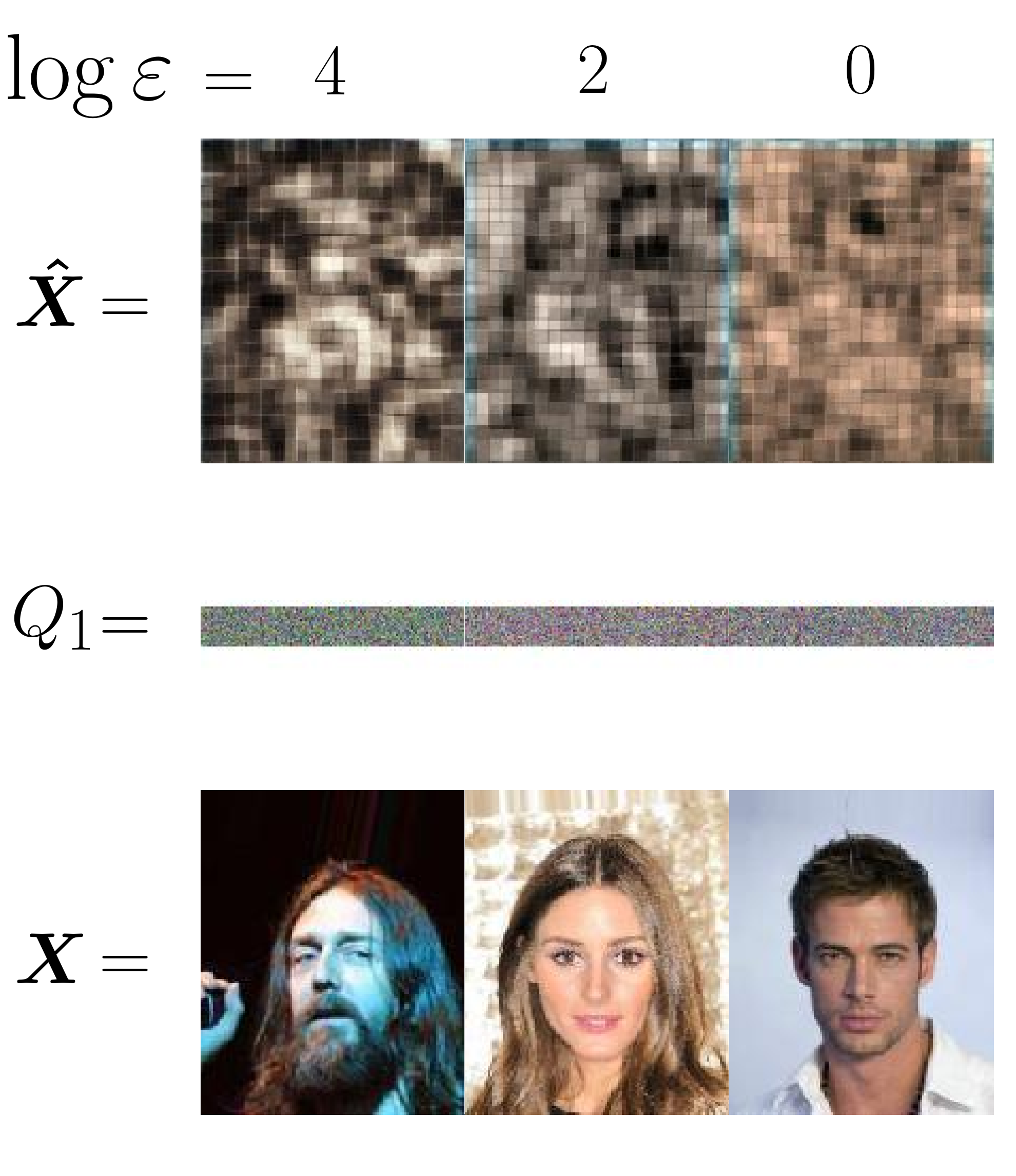}
		%\vskip -0.1in
		\caption{The image reconstruction from one of the queries in scenario two}
		\label{fig:CelebaImClient4}
	\end{subfigure}%
	\hfill
	\begin{subfigure}{0.28\textwidth}%\linewidth
		\centering
		\includegraphics[width=\linewidth,height=5.5cm,keepaspectratio]{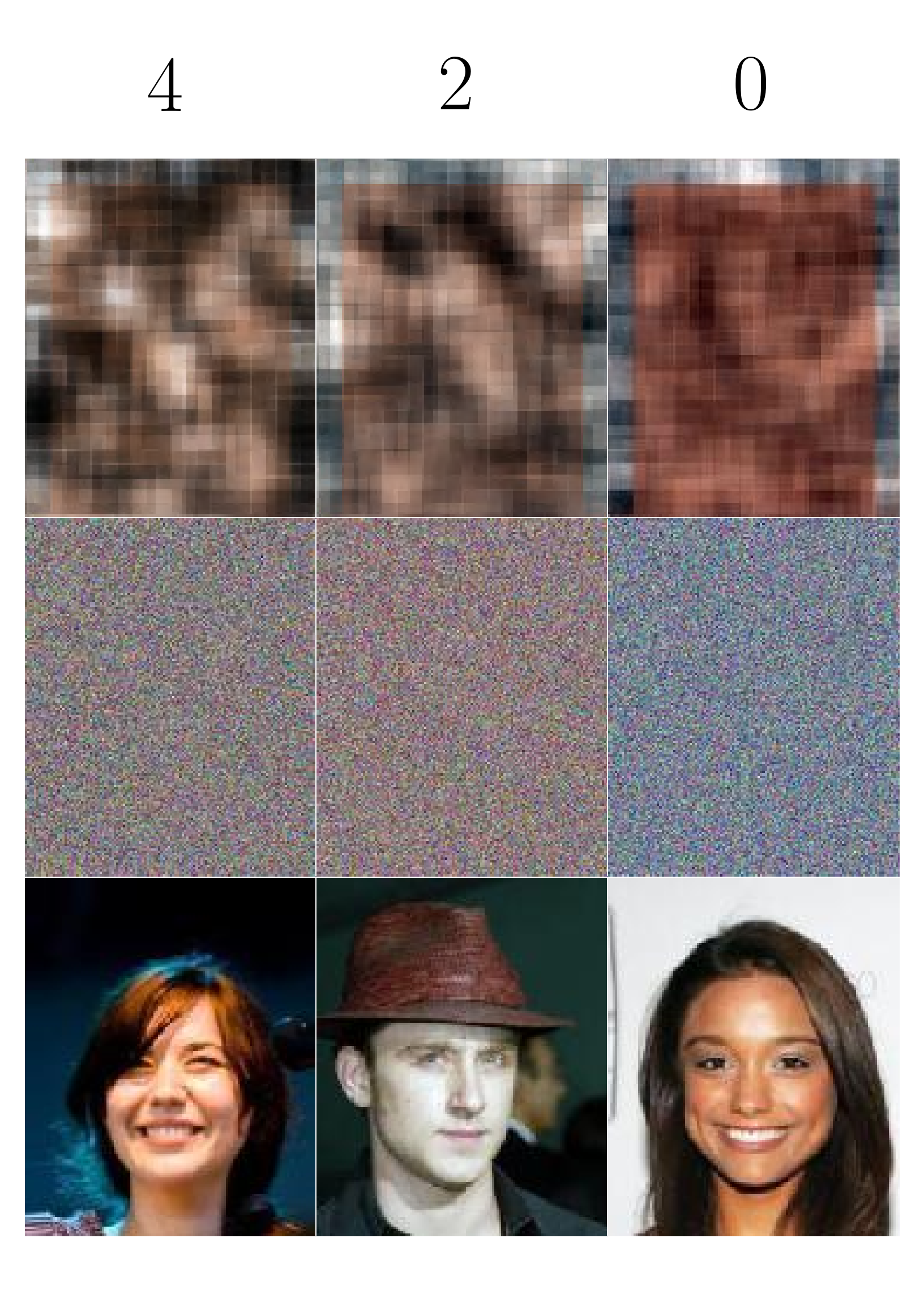}
		%\vskip -0.1in
		\caption{The image reconstruction from one of the queries in scenarios one and three}
		\label{fig:CelebaImClient5}
	\end{subfigure}%
	%\vskip -0.1in
	\caption{The image reconstruction from the query of server one. In Fig.~\ref{fig:CelebaImClient4}, one server is trained to reconstruct the image from the query of server one in scenario two, and in Fig.~\ref{fig:CelebaImClient5} from the query of server one in scenario one and three.}\label{fig:CelebaImServer}
	%\vskip -0.15in
\end{figure*}

One may ask if a single server can recover the original image from the query of one of the servers (say server one), if it is trained to. We evaluate this scenario and report in Fig.~\ref{fig:CelebaImServer}. Fig.~\ref{fig:CelebaImClient4} shows the result if we train a single server to recover the image from the query of server one in scenario two, while Fig.~\ref{fig:CelebaImClient5} shows the same result if a server is trained to recover the image from the query of server one in scenarios one and three. These figures show that reconstructing client image from the query received by one server is almost impossible.

In Table~\ref{tab:tabResults2}, we report the experimental results for various datasets, compression ratios, different scenarios, and tuples $(N,T)$. In this table, the client loss and the server loss are denoted by $l_{\textsf{Cl}}$ and $l_{\textsf{Se}}$, respectively. The server loss is equal to the loss of the image reconstruction when $T$ servers collude and run a neural network, to reconstruct the client image from their queries. The network structure is the same as what described in Table~\ref{tab:tab_encdec}, where the $T$ received queries are concatenated in the input. This means that we set the number of the input channels ($c_0$ in Scenario 2 and $c_i$ in Scenario 1 and Scenario 3)  equal to $T$ times the number of channels of a single query. These results show that the difference between the client loss and the server loss is significant. Therefore, while $T$ colluding servers learn nothing from the received queries, the client can recover the image with good quality.

\begin{table*}[!t]%icv:%[t]
	\vskip 0.3in
	\caption{The client loss and the ratio of the server loss to the client loss for the three scenarios of the offloading autoencoder tasks, compression ratio of $R_\Comp=4$ and $8$, and $(N,T)=(2,1)$ and $(3,2)$ for $\log \varepsilon=0$. (The results of Experiment 4).}
	\label{tab:tabResults2}
	%\vskip -0.1in
	\begin{center}
		\begin{small}
			%\begin{sc}
			\centering
			\begin{tabular}{lcccccccc} 
				\toprule
				\multicolumn{1}{c}{\multirow{2}{*}{Dataset}} & \multirow{2}{*}{$(N,T)$ } & \multirow{2}{*}{$R_\Comp$ } & \multicolumn{2}{c}{Scenario 1}                      & \multicolumn{2}{c}{Scenario 2}                      & \multicolumn{2}{c}{Scenario 3}                       \\ 
				\cmidrule(lr){4-5}\cmidrule(lr){6-7}\cmidrule(lr){8-9}
				\multicolumn{1}{c}{}                         &                           &                             & $l_\textsf{Cl}$  & $l_\textsf{Se} / l_\textsf{Cl}$  & $l_\textsf{Cl}$  & $l_\textsf{Se} / l_\textsf{Cl}$  & $l_\textsf{Cl}$  & $l_\textsf{Se} / l_\textsf{Cl}$   \\ 
				\toprule
				\multirow{4}{*}{MNIST}                       & \multirow{2}{*}{(2,1)}    & 4                           & 0.00219          & 22.9                             & 0.00215          & 25.3                             & 0.00536          & 9.4                               \\
				&                           & 8                           & 0.00466          & 10.8                             & 0.00494          & 11.0                             & 0.00947          & 5.3                               \\ 
				\cdashline{2-9}[1pt/1pt]
				& \multirow{2}{*}{(3,2)}    & 4                           & 0.00324          & 15.5                             & 0.00312          & 17.5                             & 0.00755          & 6.7                               \\
				&                           & 8                           & 0.00477          & 10.5                             & 0.00624          & 8.7                              & 0.01240          & 4.1                               \\ 
				\midrule
				\multirow{4}{*}{Fashion-MNIST}               & \multirow{2}{*}{(2,1)}    & 4                           & 0.00402          & 14.8                             & 0.00521          & 12.2                             & 0.00930          & 6.4                               \\
				&                           & 8                           & 0.00571          & 10.4                             & 0.00784          & 8.1                              & 0.01175          & 5.1                               \\ 
				\cdashline{2-9}[1pt/1pt]
				& \multirow{2}{*}{(3,2)}    & 4                           & 0.00451          & 13.3                             & 0.00604          & 10.4                             & 0.01135          & 5.3                               \\
				&                           & 8                           & 0.00617          & 9.7                              & 0.00907          & 7.0                              & 0.01359          & 4.4                               \\ 
				\midrule
				\multirow{4}{*}{Cifar10}                     & \multirow{2}{*}{(2,1)}    & 4                           & 0.00181          & 31.0                             & 0.00366          & 15.3                             & 0.00615          & 9.1                               \\
				&                           & 8                           & 0.00255          & 22.0                             & 0.00435          & 12.8                             & 0.00667          & 8.4                               \\ 
				\cdashline{2-9}[1pt/1pt]
				& \multirow{2}{*}{(3,2)}    & 4                           & 0.00305          & 18.3                             & 0.00550          & 10.2                             & 0.00809          & 6.9                               \\
				&                           & 8                           & 0.00356          & 15.7                             & 0.00580          & 9.7                              & 0.00801          & 7.0                               \\ 
				\midrule
				\multirow{4}{*}{CelebA}                      & \multirow{2}{*}{(2,1)}    & 4                           & 0.00138          & 61.8                             & 0.00820          & 10.4                             & 0.00969          & 8.8                               \\
				&                           & 8                           & 0.00164          & 52.0                             & 0.00928          & 9.3                              & 0.01012          & 8.4                               \\ 
				\cdashline{2-9}[1pt/1pt]
				& \multirow{2}{*}{(3,2)}    & 4                           & 0.00169          & 50.4                             & 0.00950          & 9.1                              & 0.01010          & 8.4                               \\
				&                           & 8                           & 0.00180          & 47.4                             & 0.01252          & 6.9                              & 0.01309          & 6.5                               \\
				\bottomrule
			\end{tabular}
			%\end{sc}
		\end{small}
	\end{center}
	\vskip 0.1in                           	
\end{table*}

\section{Conclusion}\label{ConclusionSection}

In this paper, we propose Corella as a privacy-preserving framework for offloading machine learning algorithms. This is particularly important for mobile devices that have limited computing resources.  In Corella, we add strong noise to the data before sending it to each server. However, to provide an opportunity to mitigate the resulting confusion and so guarantee a reasonable accuracy, the added noises to different servers are designed to be correlated.  Unlike MPC that guarantees that added noises will be eliminated completely through carefully designed excessive interaction among the servers (or alternatively using many servers), in Corella, the servers are trained to mitigate the effect of noise. Interestingly, with this method, we simultaneously achieve a high level of accuracy and a high level of privacy, with only a few servers (e.g., just two), and without any communication among the servers. Indeed, in Corella framework, in the process of training the model to predict the result (e.g., label) accurately, the model is also trained to mitigate the effect of the noises. Correlation among the noises provides the chance for this training to be successful, which is confirmed experimentally.

\pagebreak

\ifCLASSOPTIONcaptionsoff
  \newpage
\fi

%%% new %%%
\bibliographystyle{IEEEtran}
\bibliography{myref}
%%% %%% %%%

\end{document}